\documentclass{article}

\usepackage{arxiv}

\usepackage[utf8]{inputenc} % allow utf-8 input
\usepackage[T1]{fontenc}    % use 8-bit T1 fonts
\usepackage{hyperref}       % hyperlinks
\usepackage{url}            % simple URL typesetting
\usepackage{booktabs}       % professional-quality tables
\usepackage{amsfonts}       % blackboard math symbols
\usepackage{nicefrac}       % compact symbols for 1/2, etc.
\usepackage{microtype}      % microtypography
\usepackage{lipsum}
\usepackage{graphicx}
\graphicspath{ {./images/} }

\usepackage[numbers]{natbib}
\usepackage{graphicx}
\usepackage{amsmath}
\usepackage{amssymb}
\usepackage{algorithm}
\usepackage{algorithmic}
\usepackage{pifont}% 
\newcommand{\cmark}{\ding{51}}%
\newcommand{\xmark}{\ding{55}}%
\usepackage{adjustbox}
\usepackage{multirow}
\usepackage[table,xcdraw]{xcolor}
\usepackage{colortbl}
\usepackage[normalem]{ulem}
\useunder{\uline}{\ul}{}

\usepackage{times}  % DO NOT CHANGE THIS
\usepackage{helvet}  % DO NOT CHANGE THIS
\usepackage{courier}  % DO NOT CHANGE THIS

\usepackage{caption} % DO NOT CHANGE THIS AND DO NOT ADD ANY OPTIONS TO IT
\usepackage{appendix}
\graphicspath{ {./figures/} }
\usepackage{subfig}

\usepackage{newfloat}
\usepackage{listings}

\title{\textsc{GPO-VAE}: Modeling Explainable Gene Perturbation Responses utilizing GRN-Aligned Parameter Optimization}

\author{
Seungheun Baek\thanks{Equal Contributors} \\
  Department of Computer Science\\
  Korea University\\
  Seoul, South Korea\\
  \texttt{sheunbaek@korea.ac.kr} \\
   \And
 Soyon Park\footnotemark[1] \\
  Department of Computer Science\\
  Korea University\\
  Seoul, South Korea\\
  \texttt{soyon\_park@korea.ac.kr} \\
  \And
 Yan Ting Chok \\
  Department of Computer Science\\
  Korea University\\
  Seoul, South Korea\\
  \texttt{yanting1412@korea.ac.kr} \\
  \And
 Mogan Gim\thanks{Corresponding Authors} \\
  Department of Biomedical Engineering\\
  Hankuk University of Foreign Studies\\
  Yongin, South Korea\\
  \texttt{gimmogan@hufs.ac.kr} \\
  \And
 Jaewoo Kang\footnotemark[2] \\
  Department of Computer Science\\
  Korea University\\
  Seoul, South Korea\\
  \texttt{kangj@korea.ac.kr} \\
}

\begin{document}

\maketitle
\def\modelname{\textsc{GPO-VAE}}

\begin{abstract}
\textbf{Motivation:} Predicting cellular responses to genetic perturbations is essential for understanding biological systems and developing targeted therapeutic strategies. While variational autoencoders (VAEs) have shown promise in modeling perturbation responses, their limited explainability poses a significant challenge, as the learned features often lack clear biological meaning. Nevertheless, model explainability is one of the most important aspects in the realm of biological AI. One of the most effective ways to achieve explainability is incorporating the concept of gene regulatory networks (GRNs) in designing deep learning models such as VAEs. GRNs elicit the underlying causal relationships between genes and are capable of explaining the transcriptional responses caused by genetic perturbation treatments.

\textbf{Results:} We propose GPO-VAE, an explainable \textbf{VAE} enhanced by \textbf{G}RN-aligned \textbf{P}arameter \textbf{O}ptimization that explicitly models gene regulatory networks in the latent space. Our key approach is to optimize the learnable parameters related to latent perturbation effects towards GRN-aligned explainability. Experimental results on perturbation prediction show our model achieves state-of-the-art performance in predicting transcriptional responses across multiple benchmark datasets. Furthermore, additional results on evaluating the GRN inference task reveal our model's ability to generate meaningful GRNs compared to other methods. According to qualitative analysis, GPO-VAE posseses the ability to construct biologically explainable GRNs that align with experimentally validated regulatory pathways.

\textbf{Availability and Implementation:} GPO-VAE is available at \url{https://github.com/dmis-lab/GPO-VAE}

\end{abstract}

% keywords can be removed

\section{Introduction}
Predicting cellular responses is crucial for understanding biological systems and complex cellular behavior, enabling the rational manipulation of cells to develop targeted therapeutics and improve treatments for various diseases. To address the task, many computational methods have been introduced in the realm of perturbation modeling~\citep{gavriilidis2024mini}. Generative modeling approaches have emerged as a key research focus in computational gene perturbation response prediction tasks~\citep{bereket2024modelling, lopez2023learning, sohn2015learning}. In particular, variational autoencoders (VAEs) have become popular deep-learning model choices as they can learn latent features underlying perturbation responses from observed perturbation data during their training process.

However, the complex nature of perturbation experiments, including cell-specific contexts and technical variations, presents several challenges for improving the VAE's generalizability. Disentangled VAEs, with their ability to learn disentangled representations despite these complexities, have become promising tools in perturbation modeling~\citep{bereket2024modelling}. The main intuition is to enable VAEs to effectively disentangle and learn distinct latent subspaces such as those corresponding to perturbation effects and cell basal states. Notable models that pioneered this design approach are SVAE+~\citep{lopez2023learning} and SAMS-VAE~\citep{bereket2024modelling}, both of which demonstrated success in improving perturbation response predictions through disentanglement of latent subspaces. Recently, \citeauthor{baek2024cradle}~tackled the issue of technical artifacts that affect experimental perturbation data quality and introduced CRADLE-VAE, a model that additionally learns a latent subspace related to artifacts utilizing counterfactual reasoning. This approach not only enhanced the prediction accuracy and robustness of the model but also improved its generative quality.

% Gene expression profiles derived from observed single cell samples, either before or after perturbation treatments, are not only high-dimensional but also contain intricate information that may hinder the predictability of VAEs.
The main key design element in these VAEs is the Sparse Mechanism Shift hypothesis~\citep{scholkopf2021toward}. This hypothesis assumes that when a gene perturbation treatment is applied to a cell, only a few proportion of biological causal mechanisms contribute to the overall transcriptional gene expression changes. Encouraging sparsity on the latent features corresponding to gene perturbation effects has demonstrated robustness in predicting perturbation responses. 

However, explainability remains a crucial element in biological and healthcare AI, particularly from medical, legal, and ethical perspectives~\citep{amann2020explainability}. In fact, VAEs have limited explainability due to its learned latent representations lacking clear and direct associations with specific biological processes~\citep{charte2020analysis}.

One promising approach to address these explainability issues is to incorporate the concepts of gene regulatory networks (GRNs) in the core design of VAEs. GRNs provide a mechanistic framework for understanding and predicting how cells respond to genetic perturbations. By aligning the latent subspace related to gene-specific perturbation effects with gene-gene causal relationships, it is possible to retain the benefits of sparsity while enhancing model explainability. Furthermore, guiding the learning trajectories of VAEs towards formation of GRNs derived from perturbation dataset may provide valuable insights related to cell-specific biological pathways.

In this work, we introduce \modelname, a VAE that exploits the concepts of gene regulatory networks for enhanced explainability. While retaining the Sparse Mechanism Shift hypothesis as a fundamental principle, we redesign the latent subspace responsible for modeling gene perturbation effects so that it aligns with GRN topologies. Specifically, we implemented GRN-Aligned Parameter Optimization, that guides the stochastic learning of prior parameters for sampling sparsity-inducing masks towards biologically explainable GRNs. While maintaining competitive performance in predicting transcriptional responses to perturbations, we demonstrate that the GRN constructed by \modelname~not only explains the perturbation data but also effectively captures important regulatory pathways that align with experimentally validated biological pathways.

\section{Materials and Methods}

\begin{table*}[]
{\normalsize
\centering
\resizebox*{\columnwidth}{!}{%
\renewcommand{\arraystretch}{1.05}
\begin{tabular}{lllll}
\hline
\textbf{Dataset} & \textbf{\#~of  Data Instances} $\mathcal{D}$ & \textbf{\#~of All Genes} $\mathcal{G}$ & \textbf{\#~of Perturbed Genes} $\mathcal{G}^{\circ}$ & \textbf{\#~of Extended Genes} $\mathcal{G}^{+}$\\ \hline
\textbf{Replogle K562} & 129,478 & 8,563 & 622 & 924 \\ \hline
\textbf{Replogle RPE1} & 91,891 & 8,749 & 383 & 272 \\ \hline
\textbf{Adamson} & 46,236 & 17,035 & 68 & 347 \\ \hline
\end{tabular}
}
\caption{Details of each gene perturbation dataset used in this study. Note that $\mathcal{G}^{\circ},\mathcal{G}^{+}\subset\mathcal{G}$ while $\mathcal{G}^{\circ}\cap\mathcal{G}^{+}=\varnothing$. }
\label{tab:dataset_genepert}
}
\end{table*}

% 62,623 / 55
\begin{figure}[!t]
\centering
    \includegraphics[width=0.8\columnwidth]{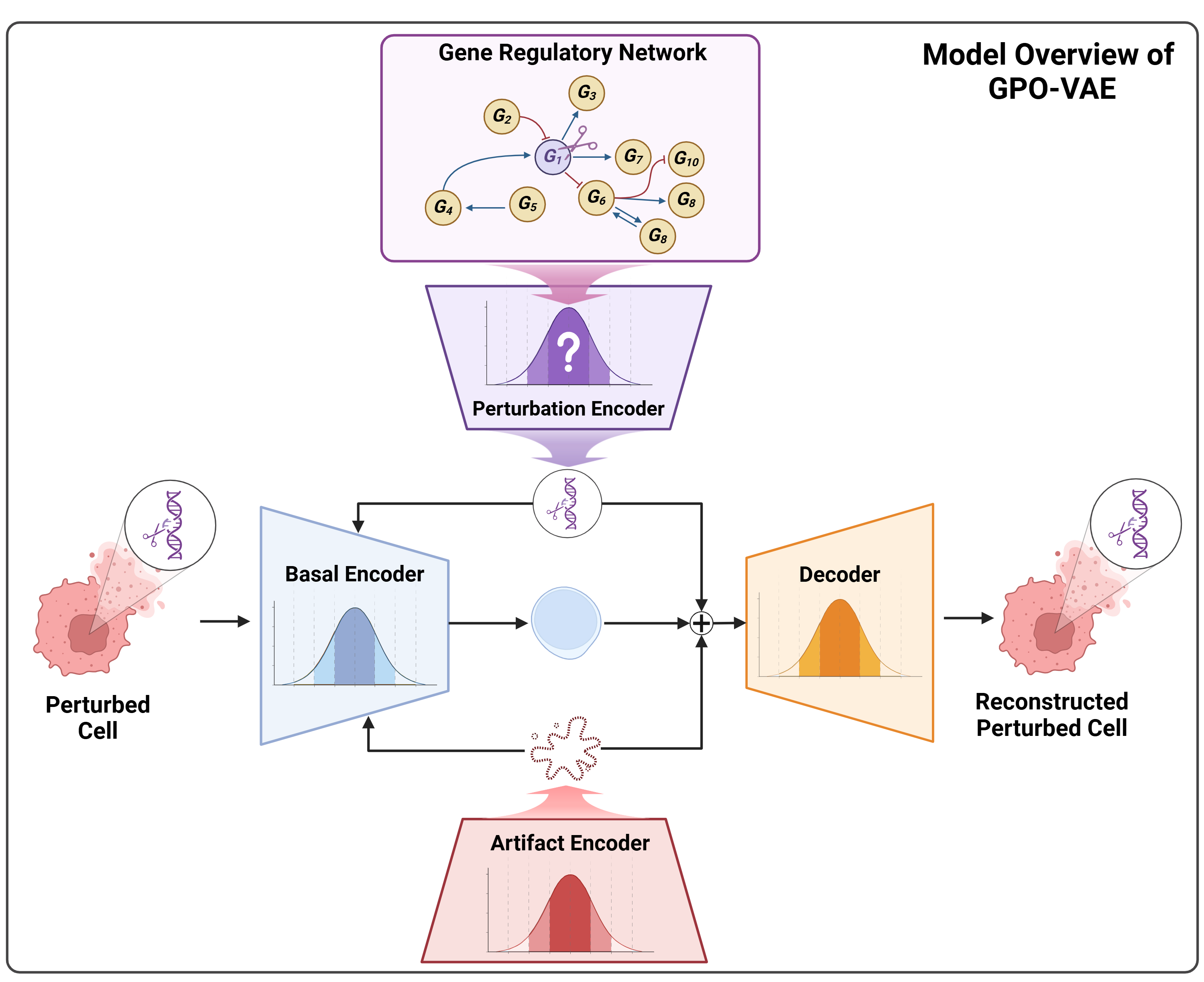}
    \caption{Model Overview of~\modelname. The model consists of three encoder modules: latent perturbation encoder, latent artifact encoder, and latent basal state encoder, and a decoder. The perturbation encoder utilizes a gene regulatory network.}
    \label{fig:model_overview}
\end{figure}

\subsection{Datasets}
We prepared three Perturb-Seq datasets (\textbf{Replogle K562} ~\citep{replogle2022mapping}, \textbf{Replogle RPE1}~\citep{replogle2022mapping} and \textbf{Adamson} ~\citep{adamson2016multiplexed}). All three datasets contain both observational (control) and interventional (post-perturbation) single-gene perturbation data instances.  The gene perturbation dataset containing $N$ data instances is a tuple consisting of $\mathcal{D}=(X,P,A)$, where $X\in\mathbb{R}^{N\times|\mathcal{G}|}$, $P\in{\{0,1\}}^{N\times\mathcal{|\mathcal{G}^{\circ}\cup\mathcal{G}^{+}|}}$, $A\in{\{0,1\}}^{N\times 1}$ correspond to the matrices of gene expression profiles, one-hot encoded gene perturbation treatments and quality control (QC) labels respectively.

Table~\ref{tab:dataset_genepert} shows the detailed gene counts of each dataset used in this study. Given the total population of genes $\mathcal{G}$ used in each dataset, where each represents a gene expression value, we denote $\mathcal{G}^{\circ}$ as the perturbation gene subset and $\mathcal{G}^{+}$ as the extended gene subset. The former contains perturbed genes for which interventional data is available, while the latter consists of additional extended genes with differential expression cutoff based on $\log FC$ and adjusted $p\text{-value}$ that were not experimentally perturbed. We expanded the gene set to increase the coverage of genetic perturbations in our model.

We adopted the preprocessing strategies from CausalBench, and further excluded samples annotated as \textit{weak perturbation} as they are likely to represent unsuccessful knockdowns, which could introduce noise and compromise the robust evaluation of GRNs~\citep{replogle2020combinatorial, chevalley2022causalbench}. As our model is an extension of CRADLE-VAE, we adopted their method for annotating the gene expression profiles with quality control (QC) pass labels using six criteria, originally stated in the filtering guidelines by Scanpy and 10X Genomics~\citep{baek2024cradle}. Details related to annotation of weak perturbation and QC labels are available in Supplementary Material \hyperref[sec:QC]{A}. 

\subsection{Model Architecture}
\modelname~is a generative model that integrates the capabilities of gene perturbation response prediction and GRN inference-based explainability. It inherits the fundamental design principles from CRADLE-VAE including latent disentanglement, sparsity assumption of perturbation effects and counterfactual reasoning. Similar to conventional VAEs, \modelname~consists of an Encoder-Decoder architecture with its encoder module composed of three submodules --- Latent Perturbation Encoder, Latent Artifact Encoder and Latent Basal State Encoder. These submodules construct latent embeddings for gene perturbation treatment, artifacts effects, and cell basal states, respectively. We modified the structure of latent features which determines the sparsity of the treatment effect so that the features represent GRNs governing the causal relations between genes. During the generative process, \modelname~takes a gene perturbation treatment vector as input and iteratively generates predicted cellular responses to compute average treatment effects. 
Figure~\ref{fig:model_overview} provides an illustrated overview of \modelname's model architecture. For clarity, we first describe the training process of each submodule, followed by the generative process.

\begin{figure*}[t]
    \includegraphics[width=\linewidth]{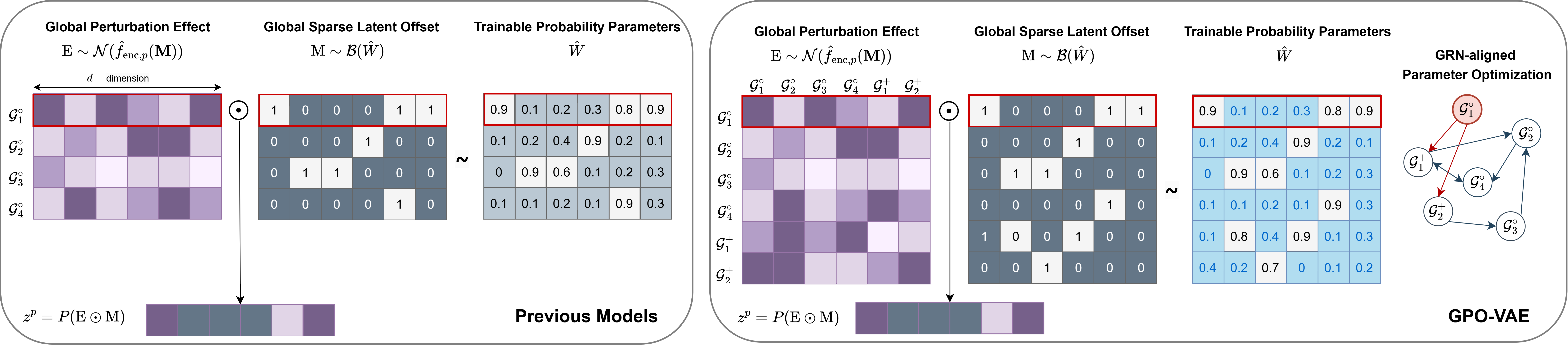}
    \caption{Comparison of perturbation encoder between previous models and our model. Unlike previous models using randomly sampled sparse latent offsets with trainable parameters, our model utilizes GRN-aligned parameter optimization for explainability.}
    \label{fig:model_pert}
\end{figure*}

\begin{table}[]
{\normalsize
\centering
\resizebox*{\columnwidth}{!}{%
\renewcommand{\arraystretch}{1.2}
\begin{tabular}{lccccc}
\hline
 & \textbf{\begin{tabular}[c]{@{}c@{}}GRN Inference\end{tabular}} &  \textbf{\begin{tabular}[c]{@{}c@{}}Artifact Disentanglement\end{tabular}} & \textbf{\begin{tabular}[c]{@{}c@{}}Basal State as Random Variable\end{tabular}} & \textbf{\begin{tabular}[c]{@{}c@{}}Sparsity Assumption\end{tabular}} \\ \hline
\textbf{Conditional VAE} & \xmark  & \xmark & \xmark & \xmark \\ \hline
\textbf{SVAE+} & \xmark & \xmark & \xmark & \cmark \\ \hline
\textbf{SAMS-VAE} & \xmark  & \cmark & \cmark & \cmark \\ \hline
\textbf{CPA-VAE} & \xmark  & \xmark & \cmark & \xmark \\ \hline
\textbf{CRADLE-VAE} & \xmark  & \cmark & \cmark & \cmark \\ \hline
\textbf{\modelname~(ours)} & \cmark & \cmark & \cmark & {\cmark} \\ \hline
\end{tabular}
}
}
\caption{List of models used in gene perturbation response prediction experiments.}
\label{tab:genepert_baselines}
\end{table}

\subsubsection{Encoder}
The Latent Perturbation Encoder aims to learn the distribution of perturbation effects, independent of biological variation and artifact effects in a cell. During training, a global sparse latent offset matrix, $\mathbf{M} \in {\left\{0,1\right\}}^{|\mathcal{G}^{\circ}\cup\mathcal{G}^{+}|\times|\mathcal{G}^{\circ}\cup\mathcal{G}^{+}|}$, is first sampled from the respective Bernoulli distribution with trainable parameters $\hat{\mathbf{W}}\in\mathbb{R}^{|\mathcal{G}^{\circ}\cup\mathcal{G}^{+}|\times|\mathcal{G}^{\circ}\cup\mathcal{G}^{+}|}$. Then, the sampled latent offsets are fed to parameterize the global latent gene-wise perturbation effect matrix, $\mathbf{E}\in\mathbb{R}^{|\mathcal{G}^{\circ}\cup\mathcal{G}^{+}|\times|\mathcal{G}^{\circ}\cup\mathcal{G}^{+}|}$, forming a Normal distribution, from which each latent perturbation effect is sampled. The latent gene perturbation effect embeddings, $\mathbf{Z}_{p}$, for each data instance are constructed by combining $\mathbf{E}$ with $\mathbf{M}$ and indexing them by the gene perturbation treatment $P$.

The Latent Perturbation Encoder that takes the gene perturbation treatments of $N$ data instances as input $P\in{\{0,1\}}^{N\times|\mathcal{G}^{\circ}\cup\mathcal{G}^{+}|}$ is mathematically expressed as follows:
\begin{align}
    \mathbf{M} &\sim \mathcal{B}(\hat{W})\label{eq:perturbation1} \\
    \mathbf{E} &\sim \mathcal{N}(\hat{f}_{\text{enc},p}(\mathbf{M}))\label{eq:perturbation2} \\
    \mathbf{Z}_{p} &= P (\mathbf{E} \odot \mathbf{M})\label{eq:perturbation3}
\end{align}

where $\mathcal{B}(\cdot)$ and $\mathcal{N(\cdot)}$ represents Bernoulli distribution and Normal distribution, respectively. $\hat{f}_{\text{enc},p}$ denotes a trainable neural network which parameterizes the distribution from which $\mathbf{M}$ is sampled.

Leveraging counterfactual reasoning, we train the Latent Artifact Encoder to target quality issues through reinforcing the disentanglement of latent variables associated with the presence of technical artifacts. The encoder takes one-hot QC labels, $A\in{\{0,1\}}^{N\times 1}$, as input and models the distribution of technical artifacts independently of basal cell features and perturbations as a Normal distribution: 
\begin{align}
\mathbf{Z}_{a} = A\mathbf{u}, \quad \text{where } \mathbf{u} \sim \mathcal{N}(\hat{\mu}, \hat{\sigma}),
\end{align}

where $\mathbf{u}$ represents the global latent artifact embedding sampled from a parameterized Normal distribution, and $\mathbf{Z}_{a} \in\mathbb{R}^{N\times d}$ denotes the latent artifact embedding matrix for $N$ data instances. As this is not the primary focus of our work, further details, including the counterfactual reasoning-based artifact disentanglement, are in the Supplementary Material \hyperref[sec:counterfactual]{B.1}.

The Latent Basal State Encoder builds latent basal state embeddings $\mathbf{Z}_{b}\in\mathbb{R}^{N\times d}$ based on integration of input gene expression profiles $\mathbf{X}$, latent perturbation effect embeddings $\mathbf{Z}_p$ and latent artifact embeddings $\mathbf{Z}_{a}$. 
\begin{align}
\mathbf{Z}_{b} \sim \mathcal{N}(\hat{f}_{\text{enc},b}(\mathbf{X},\mathbf{Z}_{p},\mathbf{Z}_{a})),  
\end{align}

where $\hat{f}_{\text{enc},b}$ denotes a trainable neural network which parameterizes the distribution from which $\mathbf{Z}_{b}$ is sampled.

\subsubsection{Decoder}
In decoding, we integrate the distinct latent embedding matrices $\mathbf{Z}_{b}$, $\mathbf{Z}_{p}$, and $\mathbf{Z}_{a}$, to reconstruct the input gene expression profile $\Tilde{\mathbf{X}}$. The reconstruction output $\Tilde{\mathbf{X}}\in\mathbb{R}^{N\times|\mathcal{G}|}$ is sampled from a neural network-equipped Gamma-Poisson distribution. The decoding process is expressed as follows: 
\begin{align}
\Lambda &\sim \Gamma(\hat{f}_{\text{dec}}(\mathbf{Z}_{b},\mathbf{Z}_{p},\mathbf{Z}_{a})L, \Theta) \\
\Tilde{\mathbf{X}} &\sim \mathcal{P}(\Lambda)
\end{align}

where $\Gamma(.)$ is the Gamma Distribution sampler, $\hat{f}_{\text{dec}}$ denotes the decoder-specialized neural network, $\mathcal{P}(.)$ is the Poisson Distribution sampler which is parameterized by $\Lambda\in\mathbb{R}^{N\times |\mathcal G|}$, $L\in\mathbb{R}^{N\times 1}$ and $\Theta\in\mathbb{R}^{N\times |\mathcal G|}$ represent the total number of read counts in each cell and trainable parameters, respectively. 

\subsubsection{Variational Inference}
Given the intractability of the data marginal likelihood $p(X|P,A)$, we introduce the correlated variational distribution $q(Z|X,P,A)$ as an approximation to the posterior distribution of the latent variables. This distribution is defined as:
\begin{align}
\begin{split}
    &q(Z_b,M,E,U|X,P,A) = \left(\prod_{t=1}^{|\mathcal{G}^{\circ}\cup\mathcal{G}^{+}|}q(\mathbf{e}_t|\mathbf{m}_t;\phi)q(\mathbf{m}_t;\phi) \right) \\
    &\quad\quad \times q(\mathbf{u};\phi)\left(\prod_{n=1}^{N} q(\mathbf{z}_{b,n}|x_n,p_n,a_n,M,E,U;\phi)\right)
\end{split}
\end{align}

To approximate the posterior distribution $\log p(X|P,A)$, we apply stochastic variational inference. The learnable parameters ($\theta$,$\phi$) of \modelname~are optimized by maximizing the evidence lower bound (ELBO), which is given by the following expression:
\begin{align}
\begin{split}
&\mathcal{J}_{rec}(\theta,\phi) = \mathbb{E}_{Z_b,M,E,U \sim q(\cdot|X,P,A;\phi)} \\
&\quad\quad\quad\quad\left[ \log \frac{p(X,Z_b,M,E,U|P,A;\theta)}{q(Z_b,M,E,U|X,P,A;\phi)} \right]
\end{split}
\end{align}

\subsubsection{Generative Process}
For generation process after training, \modelname~samples latent basal state $\mathbf{Z}_b$ from a Normal distribution $\mathcal{N}(0, I)$:
\begin{align}
\mathbf{Z}_{b} &\sim \mathcal{N}(0, I)
\end{align}

It is then concatenated with the latent artifact embedding $\mathbf{Z_a}$, and latent gene perturbation effect embeddings $\mathbf{Z}_{p}$ sampled from their respective parameterized distributions. The sampling process of $\mathbf{Z}_{p}$ is the same as \eqref{eq:perturbation1}, \eqref{eq:perturbation2}, and \eqref{eq:perturbation3}. Notably, unlike the training process, $\mathbf{A}$ is a zero matrix, representing the absence of artifacts to generate artifact-free samples:
\begin{align}
\mathbf{Z}_{a} = 0\mathbf{u}, \quad \text{where } \mathbf{u} \sim \mathcal{N}(\hat{\mu}, \hat{\sigma})
\end{align}

Finally, the concatenated embedding is passed into the trained decoder to form a Gamma-Poisson distribution, from which the reconstructed gene expression profile matrix $\Tilde{\mathbf{X}}$ is sampled.
\begin{align}
\Lambda &\sim \Gamma(\hat{f}_{\text{dec}}(\mathbf{Z}_{b},\mathbf{Z}_{p},\mathbf{Z}_{a}), \Theta,L) \\
\Tilde{\mathbf{X}} &\sim \mathcal{P}(\Lambda)
\end{align}

Formally, the joint probability distribution over the observed and latent variables is defined as:
\begin{equation}
\begin{split}
    &p(X,Z_b,M,E,U|P,A;\theta) = \left(\prod_{t=1}^{T} p(\mathbf{m}_t)p(\mathbf{e}_t)\right)p(\mathbf{u})   \\
    &\quad\quad \times \left(\prod_{n=1}^{N} p(\mathbf{z}_{b,n}) p(x_n|\mathbf{z}_{b,n},p_n,a_n,M,E,U; \theta)\right)
\end{split}
\end{equation}

\subsubsection{GRN-aligned Parameter Optimization}
As our work's core idea is modeling explainable perturbation responses, we present several modifications implemented in \modelname's Latent Perturbation Encoder. Recall that the Latent Perturbation Encoder randomly samples sparse latent offsets using its inherent Bernoulli distribution sampler with trainable parameters $\hat{W}$. In CRADLE-VAE and SAMS-VAE, $\hat{W}\in\mathbb{R}^{|\mathcal{G}^{\circ}|\times d}$, where the row vectors of the parameter matrix correspond to individual gene perturbation treatments. This design approach treats the sparse latent offsets as independent assumptions for each specific gene perturbation, enabling the model to account for distinct effects perturbation-wise.

We propose a redesigned approach for the Latent Perturbation Encoder. Specifically, we redefine $\hat{W}_{i,j}$ in $\hat{W}$ as the \textit{causal probability} from source $i$th gene to target $j$th gene. Under this modified assumption, we reshape the parameters into a square matrix $\hat{W}\in\mathbb{R}^{(|\mathcal{G}^{\circ}\cup\mathcal{G}^{+}|)\times(|\mathcal{G}^{\circ}\cup\mathcal{G}^{+}|)}$ including the extended gene subset. As illustrated in Figure~\ref{fig:model_pert}, this adjustment naturally enables interpretation of $\hat{W}$ as a gene regulatory network represented as an adjacency graph containing directed edge weights (i.e. causal gene-to-gene probabilities) and self-loops (i.e. self-regulating feedback mechanisms).

We then introduce GRN-based Parameter Optimization (GPO), a method that adjusts the parameters in $\hat{W}$ towards inclusion of topology-based model explainability to the latent perturbation effects. The loss objective from this method is composed by three components which are described as below.

First, we exploit the post-perturbational differential gene expression (DGE) values, which are calculated based on fold changes between control and perturbation gene expression profiles, as reference in adjusting the parameters. In detail, given $\hat{W}_{i,:}$ denotes the causal probabilities from source $i$th gene to all genes, each perturbation gene expression profile $\mathbf{x}^{p}_{n}$ is paired with a control gene expression profile $\mathbf{x}^{c}_{n}$ to build a reference differential gene expression profile $\Delta \mathbf{x}_n$. For the pairing process, we employed the Optimal Transport (OT) algorithm~\citep{villani2009optimal}. The reference differential gene expression profiles $\Delta\mathbf{X}\in\mathbb{R}^{N\times |\mathcal{G}^{\circ}\cup\mathcal{G}^{+}|}$ are used to guide the trainable parameters in $\hat{W}$ through a loss function, which is mathematically expressed as follows:
\begin{align}
\mathcal{J}_{dge}(\hat{W}) &= \sum||P\hat{W}-\Delta\mathbf{X}||_{1} \\
\Delta \mathbf{x}_n &=  
\begin{cases}
\text{Pairing}(\mathbf{x}^{p}_{n},\mathbf{x}^{c}_{n}) \text{ if } \mathbf{x}^{c}_{n}\neq \mathbf{0} \\
\mathbf{0} \text{ otherwise}
\end{cases}
\end{align}

Note that $\mathbf{x}^{c}_{n}$ being a zero vector indicates a control sample excluded from constructing reference differential profiles. Importantly, all reference profiles used in this optimization are derived solely from the training dataset. We denote this component as \textbf{DGE loss} ($\mathcal{J}_{dge}$).

While the above loss objective adjusts the Latent Perturbation Encoder's parameter space for its Bernoulli distribution sampler, it does not take all latent gene perturbation effects into account. In other words, only the row vectors in $\hat{W}$ which correspond to perturbation gene subset $\mathcal{G}^{\circ}$ are influenced by the loss objective, leaving the other parameters related to extended gene perturbation treatments ($\mathcal{G}^{+}$) \textit{not updated}.

Our key idea is to guide the optimization trajectories of \modelname~towards leveraging accumulated multi-hop causal relationships. As $\hat{W}$ is treated as a weighted adjacency matrix describing GRN, we can say that ${\hat{W}}^{k}$ contains $k$-hop causal relationships that explain the underlying biological mechanisms behind each gene perturbation. These $k$-hop causal relationships, which include extended genes as intermediate nodes, may contribute to expansive formation of the GRN by incrementally accumulating gene-gene relationships across different numbers of hops.

Given the reference differential gene expression profiles, we make the following modifications to $\mathcal{J}_{dge}$ as,
\begin{align}
\mathcal{J}_{dge}^{K}(\hat{W}) &= \sum||P\hat{T}_{K}-\Delta\mathbf{X}||_{1} \\
\hat{T}_{K} &= \hat{W} + \sum^{K}_{k=2}\frac{1}{|\mathcal{G^{\circ}}\cup\mathcal{G^{+}}|}\hat{W}^{k} 
\end{align}
We denote this component as \textbf{K-hop accumulation} ($\mathcal{J}_{dge}^{K}$) where $K$ is equally set to 5 across all model configurations used in our experiments.

However, the current loss objective may increase the overall magnitude of values within $\hat{T}_{K}$ as the number of hops increases. This may cause proliferation of high causal probability values, leading to higher number of predicted interactions within the GRN that may be insignificant in terms of explainability. 

To mitigate this issue, we penalize the parameters in $\hat{W}$ towards overall sparsity. Having adopted this approach from the NOTEARS algorithm~\citep{zheng2018dags}, we expect that only gene-gene interactions that are highly relevant with observed perturbation effects are retained, effectively reducing irrelevant ones. This refinement not only provides the model with more accurate information but also enhances overall predictive accuracy and explainability~\citep{xu2022sparse}.

By combining the abovementioned three components, our proposed GPO loss objective is mathematically expressed as follows,
\begin{align}
\mathcal{J}_{gpo}(\hat{W}) &= \mathcal{J}_{dge}^{K} + \mathcal{J}_{sp} \\
                           &= \sum||P\hat{T}_{K}-\Delta\mathbf{X}||_{1} + ||\hat{W}||_{1} 
\end{align}
where $\mathcal{J}_{sp}$ is the added component denoted as \textbf{sparsity penalty}.
% Figure illustrates the effects of each loss term 
The total loss objective for training \modelname~comprises maximization of the evidence lower bound (ELBO) ($\mathcal{J}_{rec}$, reconstruction of gene expression profiles), minimization of artifact disentanglement loss ($\mathcal{J}_{ade}$) and minimization of the newly proposed GPO loss objective ($\mathcal{J}_{gpo}$). The total loss objective ($\mathcal{J}$) is mathematically expressed as follows,
\begin{align}
    \mathcal{J}(\phi,\theta) = \mathcal{J}_{rec}(\theta,\phi) + \alpha\mathcal{J}_{ade}(\phi) + \beta\mathcal{J}_{gpo}(\hat{W})
\end{align}

where $\phi$ and $\theta$ are the trainable parameters in \modelname's Encoder and Decoder modules respectively. Note that $\hat{W}\subset\phi$ as the Bernoulli distribution sampler parameterized by $\hat{W}$ is the component of the Latent Perturbation Encoder.

\section{Experiments}
\subsection{Evaluation on Perturbation Response Prediction}
We conducted experiments to compare~\modelname's performance on the gene perturbation response prediction task with other VAE-based models as baselines. The baseline VAEs used in the experiments are Conditional VAE~\citep{sohn2015learning}, sVAE+~\citep{lopez2023learning}, SAMS-VAE~\citep{bereket2024modelling}, CPA-VAE~\citep{bereket2024modelling} and CRADLE-VAE~\citep{baek2024cradle}. Table~\ref{tab:genepert_baselines} summarizes the key characteristics of each VAE-variant including \modelname~and its ablations. Details on each baseline are available in Supplementary Material \hyperref[sec:pert_baselines]{C.1}.

The evaluation metrics used in the experiments are Average Treatment Effect Pearson Correlation (\textbf{ATE-}$\rho$), Average Treatment Effect R-Square Score (\textbf{ATE-}$R^2$)~\citep{bereket2024modelling}, and Jaccard Similarity between top 50 model-predicted and true differentially expressed genes (\textbf{Jaccard})~\citep{baek2024cradle}. The first two metrics compare differential expression values predicted by the model and the experimental data across all genes by calculating Pearson correlation coefficient and R-square respectively. Jaccard Similarity shows whether the predicted set of DEGs actually match the true set. Details on the evaluation metrics are available in Supplementary \hyperref[sec:eval]{D}.

\begin{table*}[hbt!]
{\Large
\resizebox*{\textwidth}{!}{%
\renewcommand{\arraystretch}{1.25}
\begin{tabular}{l|lll|lll|lll}
\hline
{\textbf{Dataset}}        & \multicolumn{3}{c|}{\textbf{Replogle K562}}                                                                                                & \multicolumn{3}{c|}{\textbf{Replogle RPE1}}                                                                                                & \multicolumn{3}{c}{\textbf{Adamson}}                                                                                                                                                                   \\ \hline
\multicolumn{1}{c|}{\textbf{Model / Method}} & \multicolumn{1}{c}{\textbf{ATE-$\rho$} $\uparrow$}     & \multicolumn{1}{c}{\textbf{ATE-$R^2$} $\uparrow$}          & \multicolumn{1}{c|}{\textbf{Jaccard} $\uparrow$}        & \multicolumn{1}{c}{\textbf{ATE-$\rho$} $\uparrow$}     & \multicolumn{1}{c}{\textbf{ATE-$R^2$} $\uparrow$}          & \multicolumn{1}{c|}{\textbf{Jaccard}$ \uparrow$}        & \multicolumn{1}{c}{\textbf{ATE-$\rho$} $\uparrow$}                         & \multicolumn{1}{c}{\textbf{ATE-$R^2$} $\uparrow$}                              & \multicolumn{1}{c}{\textbf{Jaccard} $\uparrow$}                             \\ \hline
Conditional VAE                              & 0.6440\small ± 0.00\large          & 0.4048\small ± 0.00\large          & 0.2692\small ± 0.00\large          & {\ul{0.6297\small ± 0.01\large}}    & {\ul{0.3926\small ± 0.01\large}}    & {\ul{0.3042\small ± 0.00\large}}    & 0.7176\small ± 0.02\large                              & 0.4983\small ± 0.02\large                              & 0.3619\small ± 0.01\large                              \\
SVAE+                                        & 0.4300\small ± 0.01\large          & 0.1176\small ± 0.02\large          & 0.1356\small ± 0.00\large          & 0.5317\small ± 0.03\large          & 0.2803\small ± 0.04\large          & 0.1820\small ± 0.01\large          & 0.6575\small ± 0.02\large                              & 0.4287\small ± 0.03\large                              & 0.2278\small ± 0.01\large                              \\
SAMS-VAE                                     & 0.2093\small ± 0.04\large          & -0.3172\small ± 0.06\large         & 0.0788\small ± 0.01\large          & 0.1851\small ± 0.04\large          & -0.0650\small ± 0.05\large         & 0.1188\small ± 0.01\large          & 0.5741\small ± 0.02\large                              & 0.3034\small ± 0.03\large                              & 0.2031\small ± 0.01\large                              \\
CPA-VAE                                      & 0.5740\small ± 0.02\large          & 0.3283\small ± 0.02\large          & 0.1960\small ± 0.01\large          & 0.3944\small ± 0.02\large          & 0.1463\small ± 0.02\large          & 0.1710\small ± 0.00\large          & 0.7585\small ± 0.01\large                              & 0.5686\small ± 0.01\large                              & 0.3240\small ± 0.01\large                              \\
CRADLE-VAE                                   & {\ul{0.7373\small ± 0.00\large}}    & {\ul{0.5358\small ± 0.01\large}}    & {\ul{0.3095\small ± 0.01\large}}    & 0.6248\small ± 0.03\large          & 0.3814\small ± 0.03\large          & 0.1947\small ± 0.02\large          & {\ul{0.8549\small ± 0.00\large}}                     & {\ul{0.7189\small ± 0.01\large}}                     & {\ul{0.4067\small ± 0.01\large}}                     \\ \hline
\textbf{\modelname}                            & {\textbf{0.7658\small ± 0.00\large}} & {\textbf{0.5699\small ± 0.01\large}} & {\textbf{0.3220\small ± 0.01\large}} & {\textbf{0.6507\small ± 0.02\large}} & {\textbf{0.4022\small ± 0.03\large}} & {\textbf{0.2159\small ± 0.02\large}} & {{\textbf{0.8639\small ± 0.00\large}}} & {{\textbf{0.7307\small ± 0.01\large}}} & {{\textbf{0.4138\small ± 0.01\large}}} \\ \hline
\end{tabular}
}
}
\captionsetup{justification=raggedright, singlelinecheck=false}
\caption{Quantitative evaluation on Replogle K562, Replogle RPE1, Adamson dataset for Gene Perturbation Response Prediction. The best performing results are highlighted in bold, while the second-best results are underlined.}
\label{tab:results_post_expression}
\end{table*}
\begin{table*}[hbt!]
\Large{
\resizebox*{\textwidth}{!}{%
\renewcommand{\arraystretch}{1.3}
\begin{tabular}{l|lll|lll|lll}
\hline
\textbf{Dataset}                             & \multicolumn{3}{c|}{\textbf{Replogle K562}}                                                                                            & \multicolumn{3}{c|}{\textbf{Replogle RPE1}}                                                                                            & \multicolumn{3}{c}{\textbf{Adamson}}                                                                                                                                                \\ \hline
{\textbf{Model / Method}} & \multicolumn{1}{c}{\textbf{$\mu$WD} $\uparrow$}            & \multicolumn{1}{c}{\textbf{FOR} $\downarrow$}             & \multicolumn{1}{c|}{\textbf{\# of Edges}} & \multicolumn{1}{c}{\textbf{$\mu$WD} $\uparrow$}            & \multicolumn{1}{c}{\textbf{FOR} $\downarrow$}             & \multicolumn{1}{c|}{\textbf{\# of Edges}} & \multicolumn{1}{c}{\textbf{$\mu$WD} $\uparrow$}                                   & \multicolumn{1}{c}{\textbf{FOR} $\downarrow$}                                    & \multicolumn{1}{c}{\textbf{\# of Edges}} \\ \hline
Random1000                                   & 0.114\normalsize ± 0.001\small          & 0.176\normalsize ± 0.013\small          & 1000.0\normalsize ± 0.0\small       & 0.096\normalsize ± 0.003\small          & 0.130\normalsize ± 0.017\small          & 1000.0\normalsize ± 0.0\small       & 0.080\normalsize ± 0.013\small                                 & 0.236\normalsize ± 0.015\small                                 & 100.0\normalsize ± 0.0\small       \\
Random10000                                  & 0.113\normalsize ± 0.001\small          & -1.000\normalsize ± 0.000\small         & 10000.0\normalsize ± 0.0\small      & 0.098\normalsize ± 0.001\small          & -1.000\normalsize ± 0.000\small         & 10000.0\normalsize ± 0.0\small      & 0.080\normalsize ± 0.004\small                                 & -1.000\normalsize ± 0.000\small                                & 1000.0\normalsize ± 0.0\small      \\ \hline
PC                                           & 0.154\normalsize ± 0.002\small          & 0.188\normalsize ± 0.003\small          & 1598.6\normalsize ± 17.3\small      & 0.166\normalsize ± 0.003\small          & 0.150\normalsize ± 0.002\small          & 701.2\normalsize ± 8.4\small        & 0.115\normalsize ± 0.005\small                                 & 0.228\normalsize ± 0.022\small                                 & 210.0\normalsize ± 0.0\small       \\
GES                                          & 0.160\normalsize ± 0.007\small          & 0.189\normalsize ± 0.003\small          & 636.8\normalsize ± 20.0\small       & 0.148\normalsize ± 0.009\small          & 0.153\normalsize ± 0.002\small          & 338.2\normalsize ± 9.3\small        & 0.134\normalsize ± 0.001\small                                 & 0.231\normalsize ± 0.019\small                                 & 88.0\normalsize ± 0.0\small        \\
Sortnregress                                 & 0.161\normalsize ± 0.000\small          & 0.172\normalsize ± 0.000\small          & 5998.0\normalsize ± 0.0\small       & 0.159\normalsize ± 0.006\small          & 0.111\normalsize ± 0.002\small          & 2616.8\normalsize ± 47.4\small      & 0.127\normalsize ± 0.000\small                                 & 0.204\normalsize ± 0.000\small                                 & 236.0\normalsize ± 0.0\small       \\
GRNBoost             & 0.132\normalsize ± 0.000\small          & {\ul {0.134\normalsize ± 0.000\small}}    & 150372.6\normalsize ± 829.8\small   & 0.123\normalsize ± 0.000\small          & {\ul {0.100\normalsize ± 0.000\small}}    & 59818.2\normalsize ± 438.0\small    & 0.086\normalsize ± 0.001\small                                 & {\ul {0.154\normalsize ± 0.007\small}}                           & 2288.6\normalsize ± 60.8\small     \\
NOTEARS (Linear)                             & 0.164\normalsize ± 0.000\small          & 0.192\normalsize ± 0.000\small          & 8.0\normalsize ± 0.0\small          & 0.164\normalsize ± 0.000\small          & 0.147\normalsize ± 0.015\small          & 8.0\normalsize ± 0.0\small          & 0.000\normalsize ± 0.000\small                                 & 0.240\normalsize ± 0.020\small                                 & 0.0\normalsize ± 0.0\small         \\
NOTEARS (Linear,L1)                          & 0.140\normalsize ± 0.000\small          & 0.192\normalsize ± 0.000\small          & 6.0\normalsize ± 0.0\small          & 0.164\normalsize ± 0.000\small          & 0.154\normalsize ± 0.000\small          & 8.0\normalsize ± 0.0\small          & 0.000\normalsize ± 0.000\small                                 & 0.240\normalsize ± 0.020\small                                 & 0.0\normalsize ± 0.0\small         \\ \hline
DCDI-DSF                                     & 0.162\normalsize ± 0.001\small          & 0.185\normalsize ± 0.002\small          & 4016.6\normalsize ± 38.2\small      & 0.166\normalsize ± 0.002\small          & 0.137\normalsize ± 0.003\small          & 1897.4\normalsize ± 97.1\small      & 0.110\normalsize ± 0.002\small                                 & 0.167\normalsize ± 0.003\small                                 & 537.8\normalsize ± 29.0\small      \\
DCDI-G                                       & {\ul {0.184\normalsize ± 0.002\small}}    & 0.186\normalsize ± 0.004\small          & 2125.2\normalsize ± 39.7\small      & {\ul {0.182\normalsize ± 0.008\small}}    & 0.140\normalsize ± 0.003\small          & 946.8\normalsize ± 36.0\small       & {\ul {0.138\normalsize ± 0.001\small}}                           & 0.211\normalsize ± 0.003\small                                 & 139.4\normalsize ± 8.0\small       \\
GIES                                         & 0.155\normalsize ± 0.002\small          & 0.188\normalsize ± 0.003\small          & 1547.0\normalsize ± 36.5\small      & 0.144\normalsize ± 0.003\small          & 0.151\normalsize ± 0.003\small          & 717.4\normalsize ± 82.1\small       & 0.104\normalsize ± 0.011\small                                 & 0.236\normalsize ± 0.022\small                                 & 77.0\normalsize ± 6.0\small        \\
DCDFG-MLP            & 0.137\normalsize ± 0.005\small          & 0.271\normalsize ± 0.184\small          & 9156.2\normalsize ± 2946.9\small    & 0.129\normalsize ± 0.012\small          & 0.136\normalsize ± 0.007\small          & 8696.8\normalsize ± 4004.9\small    & 0.079\normalsize ± 0.011\small                                 & 0.251\normalsize ± 0.019\small                                 & 494.4\normalsize ± 346.8\small     \\ \hline
\textbf{\modelname}                            & \textbf{0.248\normalsize ± 0.012\small} & \textbf{0.022\normalsize ± 0.007\small} & 2887.8\normalsize ± 64.9\small      & \textbf{0.280\normalsize ± 0.010\small} & \textbf{0.042\normalsize ± 0.008\small} & 2211.4\normalsize ± 217.9\small     & {\textbf{0.159\normalsize ± 0.018\small}} & {\textbf{0.014\normalsize ± 0.005\small}} & 11.0\normalsize ± 2.6\small         \\ \hline
\end{tabular}
}
}
\captionsetup{justification=raggedright, singlelinecheck=false}
\caption{Quantitative evaluation on Replogle K562, Replogle RPE1, Adamson dataset for GRN inference. The best performing results are highlighted in bold, while the second-best results are underlined.}
\label{tab:results_grn_inference}
\end{table*}
\subsection{Evaluation on Gene Regulatory Network Inference}
While~\modelname's primary role is predicting the transcriptional outcomes of gene perturbations, our work focuses on aligning the optimization of $\hat{W}$ in its Latent Perturbation Encoder towards GRN-based explainability. To quantitatively evaluate the GRNs inferred by~\modelname~based on its statistical explainability, we conducted additional experiments on GRN inference baseline methods. 

The evaluation procedures, metrics and baseline methods for these experiments were adopted from CausalBench, a benchmark framework for evaluating GRN inference methods on single-cell gene perturbation datasets~\citep{chevalley2022causalbench}. 
The baseline methods used in the experiments are categorized based on type of data used for training: observational models use only observational data, while interventional models use both observational and interventional data. Observational models include PC~\citep{spirtes2001causation}, GES~\citep{chickering2002optimal}, Sortnregress~\citep{reisach2021beware}, GRNBoost~\citep{aibar2017scenic}, NOTEARS (Linear), and NOTEARS(Linear, L1)~\citep{zheng2018dags}. Interventional models consist of DCDI-DSF~\citep{brouillard2020differentiable}, DCDI-G~\citep{brouillard2020differentiable}, GIES~\citep{hauser2012characterization}, and DCDFG-MLP~\citep{lopez2022large}. Details on each baseline are available in Supplementary Material \hyperref[sec:grn_baselines]{C.2}.

While the baseline methods featured in CausalBench directly generate the inferred networks as output, \modelname~requires extraction of the optimized parameters from the Bernoulli distribution sampler within its Latent Perturbation Encoder. Note that each ($i$,$j$)th element in our Bernoulli parameter matrix $\hat{W}$ denotes the causal probability from the $i$th gene to $j$th gene. Since the parameters are scalar values ranging from 0 to 1, we determine the presence of edges by applying a threshold of 0.5. For a fair comparison with the baseline methods, we exclude the extended genes $\mathcal{G}^{+}$ and include only the perturbed genes $\mathcal{G}^{\circ}$ in the evaluation.   

We adopted statistical evaluation frameworks from CausalBench to evaluate the model's capability of inferring GRNs that best align with gene perturbation data. Two evaluation metrics, Mean Wasserstein Distance ($\mu\mathbf{WD}$) and False Omission Rate (\textbf{FOR}) were used in these experiments. Due to the absence of ground truth, they rely on the assumption that a predicted gene-gene interaction A to B should align with the statistical effects measured by expression changes of gene B when perturbing gene A. Specifically, $\mu\mathbf{WD}$ measures the average strength of causal effects of the inferred edges. For edge from A to B, $\mathbf{WD}$ is computed between the empirical distribution of the expression of B in control samples and in A-perturbed samples. A high $\mu\mathbf{WD}$ indicates stronger causal effects on the target node imposed by the source node. \textbf{FOR} measures the proportion of inferred negative edges that are statistically significant false negatives. Additional details are available in Supplementary \hyperref[sec:eval]{D}.

\begin{table*}[hbt!]
{\normalsize
\resizebox*{\textwidth}{!}{%
\renewcommand{\arraystretch}{1.1}
\begin{tabular}{l|lll|lll}
\hline
\textbf{Model}                                                         & \multicolumn{1}{c}{\textbf{ATE-$\rho$} $\uparrow$}     & \multicolumn{1}{c}{\textbf{ATE-$R^2$} $\uparrow$}          & \multicolumn{1}{c|}{\textbf{Jaccard} $\uparrow$}        & \multicolumn{1}{c}{\textbf{$\mu$WD} $\uparrow$}            & \multicolumn{1}{c}{\textbf{FOR} $\downarrow$}             & \multicolumn{1}{c}{\textbf{\# of Edges}} \\ \hline
\begin{tabular}[c]{@{}l@{}}\modelname~($\mathcal{J}_{sp}$)\end{tabular}      & 0.6416\scriptsize ± 0.01\normalsize          & 0.3904\scriptsize ± 0.01\normalsize          & 0.2099\scriptsize ± 0.01\normalsize          & 0.115\scriptsize ± 0.004\normalsize          & {\ul {0.040\scriptsize ± 0.007\normalsize}}    & 5516.4\scriptsize ± 111.6\normalsize    \\
\begin{tabular}[c]{@{}l@{}}\modelname~($\mathcal{J}_{dge}$)\end{tabular}           & 0.6435\scriptsize ± 0.02\normalsize          & 0.3939\scriptsize ± 0.02\normalsize          & 0.2074\scriptsize ± 0.01\normalsize          & {\ul {0.192\scriptsize ± 0.001\normalsize}}    & {\ul {0.040\scriptsize ± 0.011\normalsize}}    & 7970.8\scriptsize ± 168.7\normalsize    \\
\begin{tabular}[c]{@{}l@{}}\modelname~($\mathcal{J}_{dge}^{K}$)\end{tabular} & \textbf{0.6593\scriptsize ± 0.01\normalsize} & {\ul {0.4146\scriptsize ± 0.01\normalsize}}    & {\ul {0.2197\scriptsize ± 0.01\normalsize}}    & 0.157\scriptsize ± 0.000\normalsize          & 0.042\scriptsize ± 0.005\normalsize          & 16034.2\scriptsize ± 315.3\normalsize   \\
\textbf{\modelname}~($\mathcal{J}_{gpo}$, ours)                                                             & {\ul {0.6584\scriptsize ± 0.02\normalsize}}    & \textbf{0.4169\scriptsize ± 0.02\normalsize} & \textbf{0.2215\scriptsize ± 0.01\normalsize} & \textbf{0.414\scriptsize ± 0.010\normalsize} & \textbf{0.039\scriptsize ± 0.007\normalsize} & 854.8\scriptsize ± 48.1\normalsize      \\ \hline
\end{tabular}
}
}
\captionsetup{justification=raggedright, singlelinecheck=false}
\caption{Results on ablation experiments of \modelname. The best performing results are highlighted in bold, while the second-best results are underlined.}
\label{tab:results_ablation}
\end{table*}

\subsection{Quantitative Experimental Results}

As shown in Table~\ref{tab:results_post_expression}, \modelname~outperforms the baseline models in predicting post-perturbation gene expression across all datasets. We attribute the performance improvement to the explicit modelling of GRN in the latent space. To further evaluate the robustness of the GRNs inferred by \modelname, we conducted additional experiments, with the results shown in Table~\ref{tab:results_grn_inference}. We demonstrate that the GRN inferred by \modelname~achieves the best $\mu\mathbf{WD}$ and \textbf{FOR}, despite a trade-off observed among other baseline models for these metrics. In fact, increasing the number of edges in the predicted GRN typically reduces \textbf{FOR} by decreasing the total negative edges, but it can lead to poorer performance in terms of $\mu\mathbf{WD}$ due to the inclusion of more predicted positive edges. Remarkably, \modelname~excels in both metrics while maintaining a sufficient number of edges compared to other models. This underscores \modelname's ability to construct a sparse, data-consistent causal graph for the GRN, which in turn enhances its capacity to predict perturbation responses while also improving model explainability.

\begin{figure*}[t]
\centering
    \includegraphics[width=\linewidth, height=3.25cm,keepaspectratio]{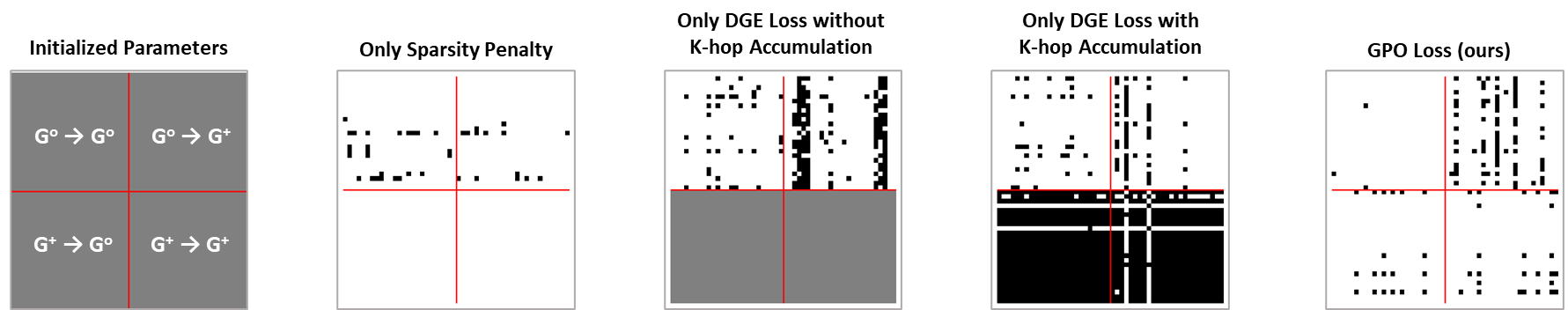}
    \caption{GRN topology analysis of GPO loss objective. Red borderlines separate perturbation and extended gene groups. The coloring scheme of the cells are based on the edge weights in each $\hat{W}$. White-colored, grey-colored and black-colored cells denote absence of edge (\textless0.5), initialized parameter (=0.5) and presence of edge (\textgreater0.5)}
    \label{fig:loss_analysis_alt}
\end{figure*}

\subsection{GRN Topology Analysis and Ablation Study}
To analyze the impact of our GRN-aligned parameter optimization, we conducted an ablation study and visualized the optimized parameters representing the directed, weighted adjacency matrix for the GRN during training in Fig~\ref{fig:loss_analysis_alt}. Note that the parameters were extracted from $\hat{W}$ within the Latent Perturbation Encoder's Bernoulli distribution sampler. Using the Replogle RPE1 dataset, we randomly sampled 25 perturbation genes and 25 extended genes for this analysis. 

The ablations for \modelname, related to the components that comprise our GPO loss objective, are the following --- only sparsity penalty ($\mathcal{J}_{sp}$), only DGE loss without K-hop accumulation ($\mathcal{J}_{dge}$) and only DGE loss with K-hop accumulation ($\mathcal{J}_{dge}^{K}$). We investigated the individual contributions of each component to the GRN-aligned optimization of $\hat{W}$.

Table~\ref{tab:results_ablation} shows the results on ablation experiment while Fig~\ref{fig:loss_analysis_alt} illustrates the inferred GRN based on different ablation settings using heatmaps. Ablation model \modelname~($\mathcal{J}_{sp}$) suffered from a huge reduction in overall statistical significance of predicted edges, leading to poor $\mu\mathbf{WD}$. Its inferred GRN exhibits edges exclusively originating from perturbation genes, with no edges involving extended genes. This outcome reflects the penalty's tendency to suppress connections related to extended genes, as their parameters were never optimized.

Conversely, ablation model \modelname~($\mathcal{J}_{dge}$) constructed a GRN with relatively less sparsity and higher $\mu\mathbf{WD}$, compared to the ablated model using only sparsity penalty. However, all inferred edges containing extended genes maintained uniform weights of 0.5 throughout model training, indicating that these relationships were never been accounted during optimization.

The GRN optimized with $\mathcal{J}_{dge}^{K}$ which has no sparsity regularization, demonstrates the involvement of both perturbed and extended genes in the optimization process. This suggests that extended genes contributing novel pathways to perturbation gene relationships were effectively optimized. However, its inferred GRN remains dense, particularly in the edges involving extended genes, decreasing $\mu\mathbf{WD}$.

Finally, the GRN inferred by our model using the GPO loss objective $\mathcal{J}_{gpo}$ is both sparse and demonstrates well-optimized relationships involving extended genes, as shown in Fig~\ref{fig:loss_analysis_alt}. The balance between the effects of sparsity penalty and multi-hop causal relationships contributed to \modelname~outperforming its ablations in terms of GRN inference along with a slight improvement in perturbation response prediction.

\begin{figure*}[t]
    \includegraphics[width=\linewidth]{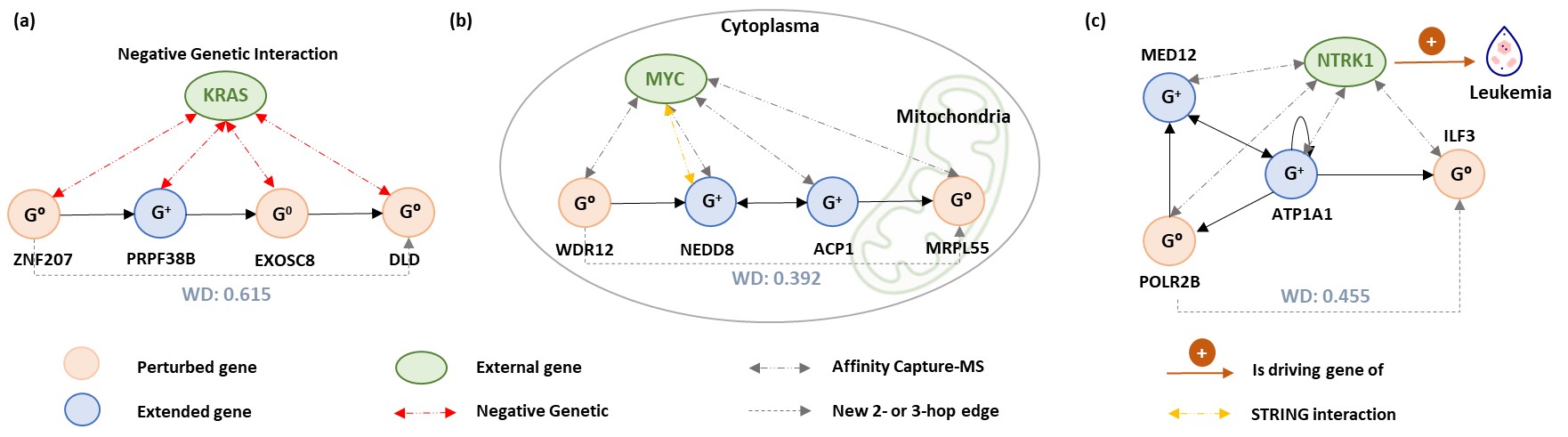}
    \caption{Case study and pathway analysis of GRN subnetworks involved in the interaction with three cancer proteins: KRAS, Myc, NTRK1. Solid lines indicate one-hop directed edges; dotted lines indicate 2- or 3-hop edges that form through the inclusion of extended genes.}
    \label{fig:qualitative analysis}
\end{figure*}

\subsection{Biological Implication of the Inferred GRN}
To investigate the impact of involving the extended gene subset in GRN inference on establishing meaningful gene-gene causal relationships, we studied the extracted subnetworks from Replogle K562 dataset, both biologically and statistically. We hypothesize that novel pathways created from the interaction of extended genes would provide biologically relevant insights related to cancer progression or immune system disorders in K562 cells, derived from a chronic myelogenous leukemia patient. Subnetworks were selected based on two criteria: 1) paths with perturbed genes connected via at least one extended gene, and 2) no directed edge between the selected perturbed genes in the GRN inferred by DCDI-G, the best baseline model. The newly formed 2- or 3-hop causal edges exhibit high $\mathbf{WD}$s, indicating strong causal effects. We then analyzed the gene sets within the subnetworks using DAVID Functional Annotation Tool.

As shown in Fig~\ref{fig:qualitative analysis}, analysis revealed that genes (or proteins) in the connected subnetworks interact with KRAS, MYC, and NTRK1, all of which are well-known cancer-associated genes (proteins), respectively. This finding highlights the consistency between the \modelname-inferred GRN and prior biological knowledge. 

In the KRAS-related subnetwork shown in Fig~\ref{fig:qualitative analysis} (a), all genes were experimentally confirmed to have a negative genetic relationship with KRAS~\citep{vichas2021integrative}. In this context, negative genetic interactions occur when mutations in individual genes cause minimal phenotypic effects, but their combined mutation in the same cell results in a severe fitness defect or lethality. Interestingly, although STRING—a widely used biological network database---did not identify interactions between these genes, experimental evidence supports the validity of our subnetwork~\citep{szklarczyk2023string}. This suggests that the inferred subnetwork could provide valuable insights for identifying new synthetic lethal gene pairs, potentially advancing target discovery efforts related to KRAS.

In Fig~\ref{fig:qualitative analysis} (b), the genes in the subnetwork is found to exhibit protein-level interaction with MYC, identified through mass spectrometric methods~\citep{solvie2022myc, wang2022ezh2, heidelberger2018proteomic}. 
Only interaction between MYC and NEDD8 is documented in STRING, highlighting the advantage of our extended subnetwork. By including extended genes such as NEDD8, which were not utilized in other baseline GRN inference methods, our approach expands the search space, enabling the discovery of additional pathways that align with prior biological networks. Moreover, leveraging the genotype-phenotype mapping derived from ~\citeauthor{replogle2022mapping}, we identified WDR12, NEDD8, and MRPL55 as genes associated with the ribosomal subunit. On the other hand, subcellular localization data from UniProtKB/Swiss-Prot revealed that MYC, WDR12, NEDD8, and ACP1 are highly likely to localize in the cytosol, while MRPL55 is localized in the mitochondrion.
% ~\citep{bateman2024uniprot}. 
Notably, the Replogle paper focused on mitochondrial genome stress responses, and prior research links MYC to cancer-related phenotypes, such as cell cycle regulation and MYC-dependent apoptosis, through mitochondrial targets. MRPL55, in particular, was reported as a nuclear-encoded mitochondrial gene downstream target in this context~\citep{morrish2014myc}. These findings suggest that the inferred MYC-related subnetwork reflects a MYC-driven pathway that transitions from the cytosol to the mitochondria, contributing to cancer-related phenotypes linked to cell cycle and mitochondrial biogenesis.

The last subnetwork shown in Fig~\ref{fig:qualitative analysis} (c) demonstrated protein-level interaction with NTRK1, an oncogenic driver discovered for leukemia studied in prior studies~\citep{joshi2019revisiting}. While STRING did not identify interactions among the genes in this subnetwork, our model revealed that MED12, ATP1A1, ILF3, and POLR2B could also serve as potential target genes for chronic myelogenous leukemia. Surprisingly, all genes in this subnetwork were associated with leukemia types, including acute myeloid leukemia, chronic lymphocytic leukemia, and chronic myelogenous leukemia as supported by prior studies~\citep{kampjarvi2014somatic, richter2022effective, lal2017somatic, nazitto2021ilf3}. This evidence suggests that the genes identified by our subnetwork could be promising candidates for discovering novel therapeutic targets for immune-related diseases.  

These results indicate that \modelname~constructs biologically and statistically meaningful edges through GRN-aligned parameter optimization of extended genes.

\subsection{Case Study on Unseen Perturbation Treatments}
We hypothesize that by leveraging the ability of the GPO loss to optimize the causal probabilities of edges formed within a k-hop neighborhood, \modelname~should be capable of predicting gene expression for unseen perturbation treatments. To verify this, we excluded samples involving three distinct perturbation treatments from the training dataset and assessed the model's performance in accurately predicting gene expression. We calculated $\mu\mathbf{WD}$ across all perturbed genes and identified the top three genes with the highest combined in-degree and out-degree, using a threshold of 0.3. These selected genes represent key hub genes within the network, as they both receive the most influence from other genes and exert the greatest influence on others. 

\begin{figure*}[t]
\centering
    \includegraphics[width=\linewidth, height=3.5cm,keepaspectratio]{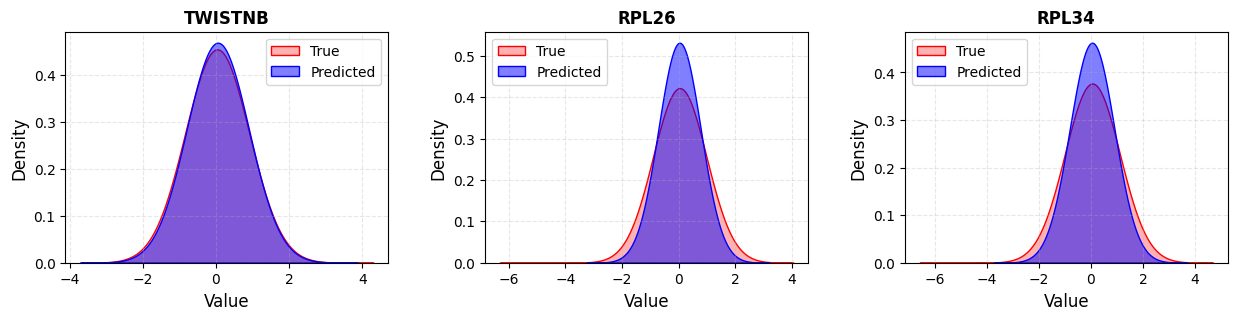}
    \caption{Quantitative performance of the model on unseen perturbations for TWISTNB, RPL26, and RPL34 genes. For TWISTNB, the ATE Pearson correlation, ATE R², and Jaccard similarity are 0.792, 0.578, and 0.587, respectively. For RPL26, the corresponding values are 0.886, 0.775, and 0.724, and for RPL34, they are 0.919, 0.831, and 0.667.}
    \label{fig:unseen}
\end{figure*}

As shown in Figure~\ref{fig:unseen}, \modelname~achieves competitive performance across all three excluded cases, with \textbf{ATE}-$\rho$ of 0.79, 0.88, and 0.91 and \textbf{ATE}-$R^2$ scores of 0.57, 0.77, and 0.83. Additionally, the predicted distributions align closely with the actual distributions, demonstrating the model’s ability to approximate post-perturbation responses. These results highlight~\modelname's ability to generalize to unseen perturbations, showcasing its robustness and scalability for gene expression prediction in practical applications. 

\section{Conclusion}
In this work, we introduce \modelname, a GRN-enhanced explainable VAE, that aligns the latent subspaces associated with gene-specific perturbation effects with gene-gene causal relationship for improved post-perturbation expression prediction. This is achieved through GRN-aligned parameter optimization, which refines the causal gene-gene relationships towards biologically meaningful GRN. Through evaluations on three benchmark datasets and case studies, we demonstrate that \modelname~not only accurately predicts cellular responses to gene perturbations but also constructs statistically robust and biologically relevant GRNs. 

One limitation of \modelname~is that, unlike perturbation response prediction models such as CRADLE-VAE and SAMS-VAE, which have demonstrated their ability to predict responses to multi-gene perturbation treatments, our model has not yet been explicitly designed for such scenarios though the architecture itself does not inherently restrict its application to multi-gene perturbation settings. 

Moreover, we identified a misalignment between the two evaluation schemes—biological and statistical—introduced in CausalBench. Furthermore, the reference networks proposed in biological evaluation schemes are not cell-type specific and may introduce biases. This observation highlights the inherent challenge of evaluating GRN inference models, as the absence of ground truth limits the comprehensiveness of existing evaluation methods. 

As future work, we aim to extend \modelname~to incorporate multi-gene perturbation treatments and enable the model to leverage relationships such as synergy, inhibition, and synthetic lethality within gene regulatory networks for improved perturbation response prediction. Also, we plan to utilize prior biological knowledge, to construct a GRN that strongly aligns with both existing biological knowledge and perturbation experimental data.

\section{Acknowledgement}
This work was supported in part by the National Research Foundation of Korea [NRF-2023R1A2C3004176], the Ministry of Health \& Welfare, Republic of Korea [HR20C0021], the Ministry of Science and ICT (MSIT) [RS-2023–00262002], and the ICT Creative Consilience Program through the Institute of Information \& Communications Technology Planning \& Evaluation(IITP) grant funded by the Korea government(MSIT)(IITP-2025-RS-2020-II201819). This work was supported by Hankuk University of Foreign Studies Research Fund (of 2024).

Figure~\ref{fig:model_overview} was created with BioRender.com.

\bibliographystyle{hunsrtnat}
% \bibliography{references}

\begin{thebibliography}{25}
\expandafter\ifx\csname natexlab\endcsname\relax\def\natexlab#1{#1}\fi
\expandafter\ifx\csname url\endcsname\relax
  \def\url#1{{\tt #1}}\fi

\bibitem{replogle2020combinatorial}Replogle, J., Norman, T., Xu, A., Hussmann, J., Chen, J., Cogan, J., Meer, E., Terry, J., Riordan, D., Srinivas, N. \& Others Combinatorial single-cell CRISPR screens by direct guide RNA capture and targeted sequencing. {\em Nature Biotechnology}. \textbf{38}, 954-961 (2020)
\bibitem{adamson2016multiplexed}Adamson, B., Norman, T., Jost, M., Cho, M., Nuñez, J., Chen, Y., Villalta, J., Gilbert, L., Horlbeck, M., Hein, M. \& Others A multiplexed single-cell CRISPR screening platform enables systematic dissection of the unfolded protein response. {\em Cell}. \textbf{167}, 1867-1882 (2016)
\bibitem{rood2024toward}Rood, J., Hupalowska, A. \& Regev, A. Toward a foundation model of causal cell and tissue biology with a Perturbation Cell and Tissue Atlas. {\em Cell}. \textbf{187}, 4520-4545 (2024)
\bibitem{chevalley2022causalbench}Chevalley, M., Roohani, Y., Mehrjou, A., Leskovec, J. \& Schwab, P. Causalbench: A large-scale benchmark for network inference from single-cell perturbation data. {\em ArXiv Preprint ArXiv:2210.17283}. (2022)
\bibitem[Baek et~al.(2024)Baek, Park, Chok, Lee, Park, Gim, and Kang]{baek2024cradle}Baek, S., Park, S., Chok, Y., Lee, J., Park, J., Gim, M. \& Kang, J. CRADLE-VAE: Enhancing Single-Cell Gene Perturbation Modeling with Counterfactual Reasoning-based Artifact Disentanglement. {\em ArXiv Preprint ArXiv:2409.05484}. (2024)
\bibitem{gavriilidis2024mini}Gavriilidis, G., Vasileiou, V., Orfanou, A., Ishaque, N. \& Psomopoulos, F. A mini-review on perturbation modelling across single-cell omic modalities. {\em Computational And Structural Biotechnology Journal}. (2024)
\bibitem{kamimoto2023dissecting}Kamimoto, K., Stringa, B., Hoffmann, C., Jindal, K., Solnica-Krezel, L. \& Morris, S. Dissecting cell identity via network inference and in silico gene perturbation. {\em Nature}. \textbf{614}, 742-751 (2023)
\bibitem{bereket2024modelling}Bereket, M. \& Karaletsos, T. Modelling cellular perturbations with the sparse additive mechanism shift variational autoencoder. {\em Advances In Neural Information Processing Systems}. \textbf{36} (2024)
\bibitem{xu2022sparse}Xu, D., Gao, E., Huang, W., Wang, M., Song, A. \& Gong, M. On the sparse DAG structure learning based on adaptive Lasso. {\em ArXiv Preprint ArXiv:2209.02946}. (2022)
\bibitem{lopez2023learning}Lopez, R., Tagasovska, N., Ra, S., Cho, K., Pritchard, J. \& Regev, A. Learning causal representations of single cells via sparse mechanism shift modeling. {\em Conference On Causal Learning And Reasoning}. pp. 662-691 (2023)
\bibitem{sohn2015learning}Sohn, K., Lee, H. \& Yan, X. Learning structured output representation using deep conditional generative models. {\em Advances In Neural Information Processing Systems}. \textbf{28} (2015)
\bibitem{spirtes2001causation}Spirtes, P., Glymour, C. \& Scheines, R. Causation, prediction, and search. (MIT press,2001)
\bibitem{chickering2002optimal}Chickering, D. Optimal structure identification with greedy search. {\em Journal Of Machine Learning Research}. \textbf{3}, 507-554 (2002)
\bibitem{zheng2018dags}Zheng, X., Aragam, B., Ravikumar, P. \& Xing, E. Dags with no tears: Continuous optimization for structure learning. {\em Advances In Neural Information Processing Systems}. \textbf{31} (2018)
\bibitem{reisach2021beware}Reisach, A., Seiler, C. \& Weichwald, S. Beware of the simulated dag! causal discovery benchmarks may be easy to game. {\em Advances In Neural Information Processing Systems}. \textbf{34} pp. 27772-27784 (2021)
\bibitem{aibar2017scenic}Aibar, S., González-Blas, C., Moerman, T., Huynh-Thu, V., Imrichova, H., Hulselmans, G., Rambow, F., Marine, J., Geurts, P., Aerts, J. \& Others SCENIC: single-cell regulatory network inference and clustering. {\em Nature Methods}. \textbf{14}, 1083-1086 (2017)
\bibitem{hauser2012characterization}Hauser, A. \& Bühlmann, P. Characterization and greedy learning of interventional Markov equivalence classes of directed acyclic graphs. {\em The Journal Of Machine Learning Research}. \textbf{13}, 2409-2464 (2012)
\bibitem{brouillard2020differentiable}Brouillard, P., Lachapelle, S., Lacoste, A., Lacoste-Julien, S. \& Drouin, A. Differentiable causal discovery from interventional data. {\em Advances In Neural Information Processing Systems}. \textbf{33} pp. 21865-21877 (2020)
\bibitem{lopez2022large}Lopez, R., Hütter, J., Pritchard, J. \& Regev, A. Large-scale differentiable causal discovery of factor graphs. {\em Advances In Neural Information Processing Systems}. \textbf{35} pp. 19290-19303 (2022)
\bibitem[Replogle et~al.(2022)Replogle, Saunders, Pogson, Hussmann, Lenail, Guna, Mascibroda, Wagner, Adelman, Lithwick-Yanai, et~al.]{replogle2022mapping}Replogle, J., Saunders, R., Pogson, A., Hussmann, J., Lenail, A., Guna, A., Mascibroda, L., Wagner, E., Adelman, K., Lithwick-Yanai, G. \& Others Mapping information-rich genotype-phenotype landscapes with genome-scale Perturb-seq. {\em Cell}. \textbf{185}, 2559-2575 (2022)
\bibitem{szklarczyk2023string}Szklarczyk, D., Kirsch, R., Koutrouli, M., Nastou, K., Mehryary, F., Hachilif, R., Gable, A., Fang, T., Doncheva, N., Pyysalo, S. \& Others The STRING database in 2023: protein–protein association networks and functional enrichment analyses for any sequenced genome of interest. {\em Nucleic Acids Research}. \textbf{51}, D638-D646 (2023)
\bibitem{tsitsiridis2023corum}Tsitsiridis, G., Steinkamp, R., Giurgiu, M., Brauner, B., Fobo, G., Frishman, G., Montrone, C. \& Ruepp, A. CORUM: the comprehensive resource of mammalian protein complexes–2022. {\em Nucleic Acids Research}. \textbf{51}, D539-D545 (2023)
\bibitem{liu2015regnetwork}Liu, Z., Wu, C., Miao, H. \& Wu, H. RegNetwork: an integrated database of transcriptional and post-transcriptional regulatory networks in human and mouse. {\em Database}. \textbf{2015} pp. bav095 (2015)
\bibitem{oughtred2021biogrid}Oughtred, R., Rust, J., Chang, C., Breitkreutz, B., Stark, C., Willems, A., Boucher, L., Leung, G., Kolas, N., Zhang, F. \& Others The BioGRID database: A comprehensive biomedical resource of curated protein, genetic, and chemical interactions. {\em Protein Science}. \textbf{30}, 187-200 (2021)
\bibitem{jang2016categorical}Jang, E., Gu, S. \& Poole, B. Categorical reparameterization with gumbel-softmax. {\em ArXiv Preprint ArXiv:1611.01144}. (2016)
\bibitem{atanackovic2023dyngfn}Atanackovic, L., Tong, A., Wang, B., Lee, L., Bengio, Y. \& Hartford, J. Dyngfn: Towards Bayesian inference of gene regulatory networks with gflownets. {\em Advances In Neural Information Processing Systems}. \textbf{36} pp. 74410-74428 (2023)
\bibitem{scholkopf2021toward}Schölkopf, B., Locatello, F., Bauer, S., Ke, N., Kalchbrenner, N., Goyal, A. \& Bengio, Y. Toward causal representation learning. {\em Proceedings Of The IEEE}. \textbf{109}, 612-634 (2021)
\bibitem{charte2020analysis}Charte, D., Charte, F., Jesus, M. \& Herrera, F. An analysis on the use of autoencoders for representation learning: Fundamentals, learning task case studies, explainability and challenges. {\em Neurocomputing}. \textbf{404} pp. 93-107 (2020)
\bibitem{amann2020explainability}Amann, J., Blasimme, A., Vayena, E., Frey, D., Madai, V. \& Consortium, P. Explainability for artificial intelligence in healthcare: a multidisciplinary perspective. {\em BMC Medical Informatics And Decision Making}. \textbf{20} pp. 1-9 (2020)
\bibitem{klys2018learning}Klys, J., Snell, J. \& Zemel, R. Learning latent subspaces in variational autoencoders. {\em Advances In Neural Information Processing Systems}. \textbf{31} (2018)
\bibitem{kulkarni2015deep}Kulkarni, T., Whitney, W., Kohli, P. \& Tenenbaum, J. Deep convolutional inverse graphics network. {\em Advances In Neural Information Processing Systems}. \textbf{28} (2015)
\bibitem{li2024learning}Li, P. \& Wu, M. Learning to refine domain knowledge for biological network inference. {\em ArXiv Preprint ArXiv:2410.14436}. (2024)
\bibitem{agrawal2024wikipathways}Agrawal, A., Balcı, H., Hanspers, K., Coort, S., Martens, M., Slenter, D., Ehrhart, F., Digles, D., Waagmeester, A., Wassink, I. \& Others WikiPathways 2024: next generation pathway database. {\em Nucleic Acids Research}. \textbf{52}, D679-D689 (2024)
\bibitem{plekha4_biogrid}BioGRID PLEKHA4.  (2024), https://thebiogrid.org/121697/summary/homo-sapiens/plekha4.html, Accessed: 2024-01-20
\bibitem{shah2019plekha4}Shah, A., Batrouni, A., Kim, D., Punyala, A., Cao, W., Han, C., Goldberg, M., Smolka, M. \& Baskin, J. PLEKHA4/kramer attenuates dishevelled ubiquitination to modulate Wnt and planar cell polarity signaling. {\em Cell Reports}. \textbf{27}, 2157-2170 (2019)
\bibitem{wikipathways_wp3888}WikiPathways WP3888: Wnt Signaling Pathway.  (2024), https://www.wikipathways.org/pathways/WP3888.html, Accessed: 2024-01-20
\bibitem{sherman2022david}Sherman, B., Hao, M., Qiu, J., Jiao, X., Baseler, M., Lane, H., Imamichi, T. \& Chang, W. DAVID: a web server for functional enrichment analysis and functional annotation of gene lists (2021 update). {\em Nucleic Acids Research}. \textbf{50}, W216-W221 (2022)
\bibitem{vichas2021integrative}Vichas, A., Riley, A., Nkinsi, N., Kamlapurkar, S., Parrish, P., Lo, A., Duke, F., Chen, J., Fung, I., Watson, J. \& Others Integrative oncogene-dependency mapping identifies RIT1 vulnerabilities and synergies in lung cancer. {\em Nature Communications}. \textbf{12}, 4789 (2021)
\bibitem{solvie2022myc}Solvie, D., Baluapuri, A., Uhl, L., Fleischhauer, D., Endres, T., Papadopoulos, D., Aziba, A., Gaballa, A., Mikicic, I., Isaakova, E. \& Others MYC multimers shield stalled replication forks from RNA polymerase. {\em Nature}. \textbf{612}, 148-155 (2022)
\bibitem{wang2022ezh2}Wang, L., Chen, C., Song, Z., Wang, H., Ye, M., Wang, D., Kang, W., Liu, H. \& Qing, G. EZH2 depletion potentiates MYC degradation inhibiting neuroblastoma and small cell carcinoma tumor formation. {\em Nature Communications}. \textbf{13}, 12 (2022)
\bibitem{heidelberger2018proteomic}Heidelberger, J., Voigt, A., Borisova, M., Petrosino, G., Ruf, S., Wagner, S. \& Beli, P. Proteomic profiling of VCP substrates links VCP to K6-linked ubiquitylation and c-Myc function. {\em EMBO Reports}. \textbf{19}, e44754 (2018)
\bibitem{binder2014compartments}Binder, J., Pletscher-Frankild, S., Tsafou, K., Stolte, C., O’Donoghue, S., Schneider, R. \& Jensen, L. COMPARTMENTS: unification and visualization of protein subcellular localization evidence. {\em Database}. \textbf{2014} (2014)
\bibitem{bateman2024uniprot}Bateman, A., Martin, M., Orchard, S., Magrane, M., Adesina, A., Ahmad, S., Bowler-Barnett, E., Bye-A-Jee, H., Carpentier, D., Denny, P. \& Others UniProt: the Universal Protein Knowledgebase in 2025. {\em Nucleic Acids Research}. (2024)
\bibitem{morrish2014myc}Morrish, F. \& Hockenbery, D. MYC and mitochondrial biogenesis. {\em Cold Spring Harbor Perspectives In Medicine}. \textbf{4}, a014225 (2014)
\bibitem{joshi2019revisiting}Joshi, S., Davare, M., Druker, B. \& Tognon, C. Revisiting NTRKs as an emerging oncogene in hematological malignancies. {\em Leukemia}. \textbf{33}, 2563-2574 (2019)
\bibitem{kampjarvi2014somatic}Kämpjärvi, K., Järvinen, T., Heikkinen, T., Ruppert, A., Senter, L., Hoag, K., Dufva, O., Kontro, M., Rassenti, L., Hertlein, E. \& Others Somatic MED12 mutations are associated with poor prognosis markers in chronic lymphocytic leukemia. {\em Oncotarget}. \textbf{6}, 1884 (2014)
\bibitem{richter2022effective}Richter, A., Lange, S., Holz, C., Brock, L., Freitag, T., Sekora, A., Knuebel, G., Krohn, S., Schwarz, R., Hinz, B. \& Others Effective tumor cell abrogation via Venetoclax-mediated BCL-2 inhibition in KMT2A-rearranged acute B-lymphoblastic leukemia. {\em Cell Death Discovery}. \textbf{8}, 302 (2022)
\bibitem{lal2017somatic}Lal, R., Lind, K., Heitzer, E., Ulz, P., Aubell, K., Kashofer, K., Middeke, J., Thiede, C., Schulz, E., Rosenberger, A. \& Others Somatic TP53 mutations characterize preleukemic stem cells in acute myeloid leukemia. {\em Blood, The Journal Of The American Society Of Hematology}. \textbf{129}, 2587-2591 (2017)
\bibitem{nazitto2021ilf3}Nazitto, R., Amon, L., Mast, F., Aitchison, J., Aderem, A., Johnson, J. \& Diercks, A. ILF3 is a negative transcriptional regulator of innate immune responses and myeloid dendritic cell maturation. {\em The Journal Of Immunology}. \textbf{206}, 2949-2965 (2021)
\bibitem{villani2009optimal}Villani, C. \& Others Optimal transport: old and new. (Springer,2009)


\end{thebibliography}

\maketitle
\def\modelname{\textsc{GPO-VAE}}

\clearpage
\onecolumn

\appendix
\renewcommand{\thesection}{\Alph{section}.\arabic{section}}
\setcounter{section}{0}

\begin{appendices}
\section{Quality Control (QC) Criteria}
\label{sec:QC}
We adopted the six QC criteria defined in CRADLE-VAE~\citep{baek2024cradle}, in line with 10X Genomics, for the QC annotation of each of the gene expression data instances. The QC criteria are as follows: 
1. UMI Counts: Unique Molecular Identifiers (UMIs) represent the distinct molecular tags added to RNA molecules during the preparation phase of single-cell RNA sequencing (scRNA-seq). The total UMI count per cell indicates the number of unique RNA molecules detected. 
Filtering out cells with very low UMI counts can help reduce noise and improve data reliability. 
2. Number of Features: This metric refers to the number of unique genes or transcripts 
identified in each cell. Filtering out cells with excessively high or low feature counts is 
essential to remove potential anomalies, such as multiplets or droplets contaminated with 
ambient RNA. A high feature count suggests that the cell is expressing a diverse set of 
genes, typically seen in viable and healthy cells. Conversely, cells with very few features are likely non-viable and are often excluded during preprocessing. 
3. Mitochondrial Read Percentage: The proportion of RNA reads originating from 
mitochondrial genes is an indicator of cellular health. Cells with a high percentage of 
mitochondrial RNA are often stressed or damaged, making this metric a valuable filter during quality control. 
4. Hemoglobin Read Percentage: In scRNA-seq, hemoglobin-associated transcripts reflect the 
activity of hemoglobin genes, primarily found in red blood cells. In experiments focusing on non-hematopoietic tissues, a high percentage of hemoglobin reads may indicate sample 
contamination or preparation issues. 
5. Ribosomal Read Percentage: This refers to the fraction of sequencing reads derived from 
ribosomal RNA (rRNA). An unusually high proportion of ribosomal reads might suggest an 
inefficient rRNA depletion step, leading to less usable data for gene expression analysis as rRNA can dominate the sequencing output. 
6. Doublets: Doublets occur when two or more cells are accidentally captured together in the same droplet or well during sequencing. These artifacts result in combined gene expression profiles that can mimic or distort real biological states, making it necessary to detect and exclude doublets to maintain data accuracy.

\section{Methods} 
\subsection{Counterfactual Reasoning} 
\label{sec:counterfactual}
Counterfactual reasoning in machine learning explores hypothetical scenarios by estimating potential outcomes under alternative conditions. This approach addresses the question: 
``What would the model predict if a different action had been taken?" In the context of causal relationships, counterfactual analysis is particularly valuable for understanding the impact of various treatments or interventions. In this study, we employ counterfactual reasoning to investigate a specific question: 
``What would the outcome have been if the result had not been influenced by technical artifacts, given a particular treatment (perturbation)?" 

Following CRADLE-VAE, we employed counterfactual reasoning-based artifact disentanglement in the latent artifact encoder module. Latent Artifact Encoder models the distribution of technical artifacts independently of basal cell features and perturbations. During training, it samples a global latent artifact embedding $\mathbf{u}\in\mathbb{R}^{1\times d}$ from a parameterized Normal distribution $\mathcal{N}(\hat \mu, \hat \sigma)$. The sampled embedding $\mathbf{Z_a}\in n\times d$, which represents a technical artifact, is then masked if the input gene expression profile is annotated as artifact-free with a QC label of 0 (QC passed), otherwise it is retained ($A\in{\{0,1\}}^{N\times 1}$). In addition, it creates a counterfactual $\mathbf{Z_{a,c}}\in n\times d$ of the global artifact embedding by applying the opposite masking operation ($1-A$) (i.e., retaining the embedding if it was originally zeroed and vice versa). 

The Latent Artifact Encoder that takes the QC labels of n data instances as input ($A\in{\{0,1\}}^{N\times 1}$) is mathematically expressed as follows, 
\begin{align}
\mathbf{u} &\sim \mathcal{N}(\hat{\mu}, \hat{\sigma}) \\
\mathbf{Z}_{a} &= A\mathbf{u} \\
\mathbf{Z}_{a,c} &= (1-A)\mathbf{u}            
\end{align}
where, 
\begin{itemize}
    \item $\mathbf{u}\in\mathbb{R}^{1\times d}$ is the global latent artifact embedding. 
    \item $\mathcal{N}(\hat{\mu}, \hat{\sigma}$) is the Normal Distribution sampler with trainable parameters $\hat{\mu}\in\mathbb{R}^{d}$, $\hat{\sigma}\in\mathbb{R}^{d}$. 
    \item $\mathbf{Z}_{a}\in{n \times d}$ is the latent artifact embedding matrix for the $n$ data instances. 
    \item $\mathbf{Z}_{a,c}\in{n \times d}$ is the counterfactual latent artifact embedding matrix for the $n$ data instances.
\end{itemize}
Meanwhile, the Latent Basal State Encoder that takes the gene expression profiles of $n$ data instances as input ($X\in{\{0,1\}}^{n\times|\mathcal{G}|}$) is mathematically expressed as follows, 
\begin{align}
\mathbf{Z}_{b,c} &\sim \mathcal{N}(\hat{f}_{(\text{enc},\mu)}(\mathbf{X},\mathbf{Z}_{p},\mathbf{Z}_{a,c}),
                                   \hat{f}_{(\text{enc},\sigma)}(\mathbf{X},\mathbf{Z}_{p},\mathbf{Z}_{a,c}))   \\
\bar{\mathbf{Z}}_{b,c} &\sim \mathcal{N}(\hat{f}_{(\text{enc},\mu)}(\bar{\mathbf{X}},\mathbf{Z}_{p},\mathbf{Z}_{a,c}),
                                         \hat{f}_{(\text{enc},\sigma)}(\bar{\mathbf{X}},\mathbf{Z}_{p},\mathbf{Z}_{a,c}))   \\
\mathbf{Z}_{b} &\sim \mathcal{N}(\hat{f}_{(\text{enc},\mu)}(\mathbf{X},\mathbf{Z}_{p},\mathbf{Z}_{a}),
                                 \hat{f}_{(\text{enc},\sigma)}(\mathbf{X},\mathbf{Z}_{p},\mathbf{Z}_{a}))  
\end{align}
 where, 
\begin{itemize}
    \item $\hat{f}_{(\text{enc},\mu)}$,$\hat{f}_{(\text{enc},\sigma)}$ are the encoder-specialized neural networks that output the mean, variance parameters for the Normal Distribution sampler $\mathcal{N}(\cdot,\cdot)$ respectively. 
    \item $\mathbf{Z}_{b,c}\in\mathbb{R}^{n\times d}$ is the counterfactual latent basal state embedding matrix for the $n$ data instances. 
    \item $\bar{\mathbf{Z}}_{b,c}\in\mathbb{R}^{n\times d}$ is the reference counterfactual latent basal state embedding matrix for the $n$ data instances. 
    \item $\mathbf{Z}_{b}\in\mathbb{R}^{n\times d}$ is the latent basal state embedding matrix for the $n$ data instances. 
\end{itemize}
 Here, an artifact disentanglement loss objective is imposed to guide $\mathbf{z}^b_{i,c}$ to align with $\bar{\mathbf{z}}^b_{i,c}$. Introduced in~\citep{baek2024cradle}, this is done by minimizing the KL divergence between the latent basal state embeddings between counterfactual latent basal state embedding and the reference counterfactual latent basal state embedding. The artifact disentanglement loss is formally defined as: 
 \begingroup
\small
\begin{align}
    \mathcal{J}_{ade}(\phi) = -\text{KL}\left[ q(Z^b_c | X, P, A; \phi) \| q(\bar{Z}^b_c | \bar{X}, P, A; \phi) \right]
\end{align} \\
\endgroup

\subsection{Optimal Transport} 
We employed optimal transport in sample pairing between interventional data observational data to calculate  for optimization of GRN using DGE loss. Optimal Transport (OT) is commonly employed to assess the similarity between distributions, particularly when their supports do not overlap. In cases where the supports are disjoint, OT-based Wasserstein distances offer advantages over widely used f-divergences, such as the Kullback-Leibler divergence, Jensen-Shannon divergence, and Total Variation distance. The objective of OT is as below:  
\begingroup
\small
\begin{align}
    \min_{m, m \# \mu_s = \mu_t} \int c(x,m(x))d\mu_s(x)
\end{align} \\
\endgroup
where $c(\cdot , \cdot)$ is the ground cost.
For implementation details, refer to \url{https://pythonot.github.io/}

\newpage
\section{Baselines} 
\subsection{VAE-based Perturbation Response Prediction Models} 
\label{sec:pert_baselines}
Below includes brief descriptions of the baselines we employed in our experiments. 

\subsubsection{Conditional VAE}  
Conditional VAE, first introduced for structured output prediction tasks, is a deep generative model that utilizes stochastic neural networks with Gaussian latent variables. It incorporates features including a VAE backbone and input omission noise, which helps regularize the deep neural network during the training process. 

\subsubsection{SVAE+} 
In line with the sparse mechanism shift hypothesis, SVAE+ tackles data sparsity explicitly through a masking and embedding mechanism. However, different from other VAE-based baseline models, the model architecture itself is not designed for modelling the treatment effect of multiple interventions. Rather, it generates a cell’s complete latent embedding by sampling from a learned prior conditioned on the treatment applied to the cell.  

\subsubsection{SAMS-VAE}
SAMS-VAE is a VAE-based generative model designed for modelling cellular perturbation effect. It incorporates sparsity in the latent global perturbation variables to learn disentangled, pertrubation-specific latent subspace and the latent basal state. It extends SVAE+ by capturing interventions and their sparse effects as explicit latent additive latent variables. 

\subsubsection{CPA-VAE}
CPA-VAE is an ablated version of SAMS-VAE defined by Bereket and Karaletsos, where all mask components are fixed to 1. In other words, it excludes sparsity from the latent perturbation effects. However, it retains the advantages of the improved inference methods used for correlated variational families. 

\subsubsection{CRADLE-VAE}
CRADLE-VAE addresses the issue of unwanted artifacts effect present in data by utilizing the quality control (QC) annotation, and thereby increasing the number of gene expression data available for training. It adopts the architecture of SAMS-VAE while additionally incorporating a latent artifact encoder for artifacts effect distribution modelling. Our model inherits the basic architecture and concepts in CRADLE-VAE but increases the explainability of the latent subspaces by aligning it to gene regulatory networks via GRN loss. 

\newpage
\subsection{Gene Regulatory Network Inference Methods}
\label{sec:grn_baselines}
We adopted the baselines for both of which leverages observational and interventional data, as introduced in CausalBench challenges.  
\subsubsection{PC} 
Though not specifically designed for gene regulatory network inference, this method is among the most commonly employed approaches for causal inference using observational data. It relies on the assumption of no confounding variables and determines conditional independence to provide results that are asymptotically accurate. The output is an equivalence class of graphs that align with the outcomes of the conditional independence tests.  
\subsubsection{Greedy Equivalence Search (GES)}  
GES is a general score-based algorithm for inferring causal structures from observational data. This method uses a two-phase process, consisting of Forward and Backward phases, to iteratively add and remove edges from the graph while computing a score to select within an equivalence class.  
\subsubsection{SortnRegress}  
Sortnregress is a machine learning-based method specifically developed for Gene Regulatory 
Network (GRN) inference, focusing on ranking and regression to identify regulatory relationships. The method evaluates candidate parent genes for each target gene by their ability to predict the expression levels of the downstream gene. 
\subsubsection{GRNBoost} 
Unlike other algorithms above, GRNBoost is a method specifically designed for Gene Regulatory Network (GRN) inference, leveraging Gradient Boosting tree models to identify regulatory relationships. GRNBoost operates by evaluating candidate parent genes for each target gene, ranking them based on their ability to predict the expression profile of the downstream gene. This ranking serves as a feature selection process, helping to identify the most relevant interactions and infer the underlying network structure. 
\subsubsection{NOTEARS}  
This approach reformulates the problem of inferring a DAG into a continuous optimization task over real-valued matrices, bypassing the need for combinatorial searches across acyclic graphs. It accomplishes this by designing a smooth function, with easily computable derivatives, that equals zero only when the corresponding graph is acyclic. Different variants of NOTEARS are defined by the type of regularization term included in the loss function (e.g., L1 regularization to enforce sparsity). 
\subsubsection{Differentiable Causal Discovery with Interventions (DCDI)} 
DCDI is a framework that combines neural networks and gradient-based optimization techniques to infer Directed Acyclic Graphs (DAGs) by leveraging interventional data. It is specifically designed to handle various types of interventions, including perfect interventions (where the target variable is completely determined), imperfect interventions (where the target variable retains some level of uncertainty), and unknown interventions (where the target of intervention is not explicitly identified). 
Variants of DCDI include DCDI-G which assumes that the conditional distributions of variables given their parents are Gaussian, DCDI-DSF which removes the Gaussian assumption and utilizes normalizing flows to model more complex and flexible conditional distribution. 
\subsubsection{Greedy Interventional Equivalence Search (GIES)}
GIES extends GES to incorporate interventional data, which provides additional information about causal relationships. Interventional data is generated when certain variables are experimentally manipulated, breaking natural dependencies and revealing causal directions. To leverage intervention data, GIES adds a third phase, called the "turning phase," which adjusts the orientation of edges based on interventional data. 
\subsubsection{Random (k)} 
Random (k) serves as the simplest baseline, generating a graph by randomly selecting k nodes uniformly without replacement. In our experiments, we evaluated the approach using k=1000. 

\newpage
\section{Evaluation Metrics} 
\label{sec:eval}
\subsection{Average Treatment Effect-Pearson (ATE-$\rho$) $\And$ -$R^2$ (ATE-$R^2$)} 
As employed by Bereket and Karaletsos, ATE-Pearson and ATE-R2 is a metric that measures the correlation between the average treatment effect of the predicted expression and the  average differential expression (DE) of the real data. In our experiments, we estimate the model average treatment effect with K = 2,500 particles for reporting the results. 
\subsection{Jaccard Top k} 
Jaccard Top k measures the ability of a model in accurately predicting the top k differentially expressed genes given a type of perturbation. Specifically, we calculate the Jaccard Index between the set of top-k differentially expressed genes predicted by the model and the one from the real data.   
\subsection{Mean Wasserstein Distance ($\mu$WD)} 
Mean Wasserstein Distance ($\mu$WD) measures the average strength of causal effects of the inferred edges. For edge from A to B, Wasserstein distance is computed between the empirical distribution of the expression of B in control samples and in A-perturbed samples. Then the average of all inferred edges is calculated. A high $\mu$WD indicates stronger causal effects on the child imposed by the parent. 
\subsection{False Omission Rate (FOR)} 
FOR measures the proportion of False Negatives among the inferred Negatives. Due to the absence of ground truth graph, false negatives are determined by performing two-sided Mann–Whitney U rank test between samples from the expression of B in control samples and in A-perturbed samples for all sampled negative pairs. An edge is deemed false negative if the null hypothesis that the two distributions are equal is rejected with a p-value threshold equal or lower than 0.05. The number of negative samples we use in our experiments is 500.

\section{Additional Results}
\begin{table*}[hbt!]
\resizebox*{\textwidth}{!}{%
\renewcommand{\arraystretch}{1.1}
\begin{tabular}{c|c|c|cc|cc|cc|cc|cc}
\hline
Network               & \begin{tabular}[c]{@{}c@{}}$\mu$WD\end{tabular} & \# of Edges          & \multicolumn{2}{c|}{\begin{tabular}[c]{@{}c@{}}Pooled \\ Biological Evaluation\end{tabular}} & \multicolumn{2}{c|}{CORUM}                                   & \multicolumn{2}{c|}{\begin{tabular}[c]{@{}c@{}}STRING-network\end{tabular}} & \multicolumn{2}{c|}{RegNetwork}                              & \multicolumn{2}{c}{BioGRID}                                \\ \hline
\multicolumn{1}{l|}{} & \multicolumn{1}{l|}{}                                          & \multicolumn{1}{l|}{} & \multicolumn{1}{l|}{precision}                   & \multicolumn{1}{l|}{recall}                  & \multicolumn{1}{l|}{precision} & \multicolumn{1}{l|}{recall} & \multicolumn{1}{l|}{precision}          & \multicolumn{1}{l|}{recall}          & \multicolumn{1}{l|}{precision} & \multicolumn{1}{l|}{recall} & \multicolumn{1}{l}{precision} & \multicolumn{1}{l}{recall} \\ \hline
GT($>$0.10)             & 0.170                                                          & 59293                 & \multicolumn{1}{c|}{0.289}                       & 0.177                                        & \multicolumn{1}{c|}{0.031}     & 0.216                       & \multicolumn{1}{c|}{0.281}              & 0.179                                & \multicolumn{1}{c|}{0.002}     & 0.309                       & 0.041                         & 0.193                      \\
GT($>$0.18)             & 0.270                                                          & 16908                 & \multicolumn{1}{c|}{0.301}                       & 0.053                                        & \multicolumn{1}{c|}{0.037}     & 0.074                       & \multicolumn{1}{c|}{0.293}              & 0.053                                & \multicolumn{1}{c|}{0.002}     & 0.088                       & 0.046                         & 0.062                      \\
GT($>$0.25)             & 0.358                                                          & 6956                  & \multicolumn{1}{c|}{0.294}                       & 0.021                                        & \multicolumn{1}{c|}{0.040}     & 0.033                       & \multicolumn{1}{c|}{0.287}              & 0.021                                & \multicolumn{1}{c|}{0.002}     & 0.034                       & 0.044                         & 0.024                      \\
Pool                  & 0.065                                                          & 96454                 & \multicolumn{1}{c|}{1.000}                       & 1.000                                        & \multicolumn{1}{c|}{0.088}     & 1.000                       & \multicolumn{1}{c|}{0.963}              & 1.000                                & \multicolumn{1}{c|}{0.002}     & 0.500                       & 0.130                         & 1.000                      \\
CORUM                 & 0.073                                                          & 8528                  & \multicolumn{1}{c|}{1.000}                       & 0.088                                        & \multicolumn{1}{c|}{1.000}     & 1.000                       & \multicolumn{1}{c|}{0.940}              & 0.086                                & \multicolumn{1}{c|}{0.013}     & 0.239                       & 0.458                         & 0.312                      \\
STRING                & 0.066                                                          & 92886                 & \multicolumn{1}{c|}{1.000}                       & 0.963                                        & \multicolumn{1}{c|}{0.086}     & 0.940                       & \multicolumn{1}{c|}{1.000}              & 1.000                                & \multicolumn{1}{c|}{0.002}     & 0.462                       & 0.101                         & 0.753                      \\
RegNetwork            & 0.085                                                          & 476                   & \multicolumn{1}{c|}{0.500}                       & 0.002                                        & \multicolumn{1}{c|}{0.239}     & 0.013                       & \multicolumn{1}{c|}{0.462}              & 0.002                                & \multicolumn{1}{c|}{1.000}     & 1.000                       & 0.349                         & 0.013                      \\
BioGRID               & 0.068                                                          & 12503                 & \multicolumn{1}{c|}{1.000}                       & 0.130                                        & \multicolumn{1}{c|}{0.312}     & 0.458                       & \multicolumn{1}{c|}{0.753}              & 0.101                                & \multicolumn{1}{c|}{0.013}     & 0.349                       & 1.000                         & 1.000                      \\ \hline
\end{tabular}
}
\end{table*}
\subsection{Misalignment between biological evaluation and statistical evaluation results}  
The results below demonstrate a misalignment of biological evaluation and statistical evaluation of biological networks. GT represents the ground truth wasserstein distance-based network by applying a threshold of 0.10, 0.18, and 0.25, respectively. Biological networks included in the evaluation are as follows: 
\begin{itemize}
    \item CORUM : \url{https://mips.helmholtz-muenchen.de/corum/}
    \item STRING : \url{https://version11.string-db.org/} 
    \item RegNetwork : \url{https://regnetworkweb.org/}
    \item BioGRID : \url{https://thebiogrid.org/}
    \item ‘Pool’ represents the union of all biological networks stated above 
\end{itemize}

\subsection{GRN toplogy analysis}
The figure below shows the GRN topology analysis of GPO loss objectives in terms of edge weight (causal probability) where darker orange indicates higher probability and lighter orange indicates lower probability: 

\begin{figure*}[t]
\centering
    \includegraphics[width=\linewidth, height=3.25cm,keepaspectratio]{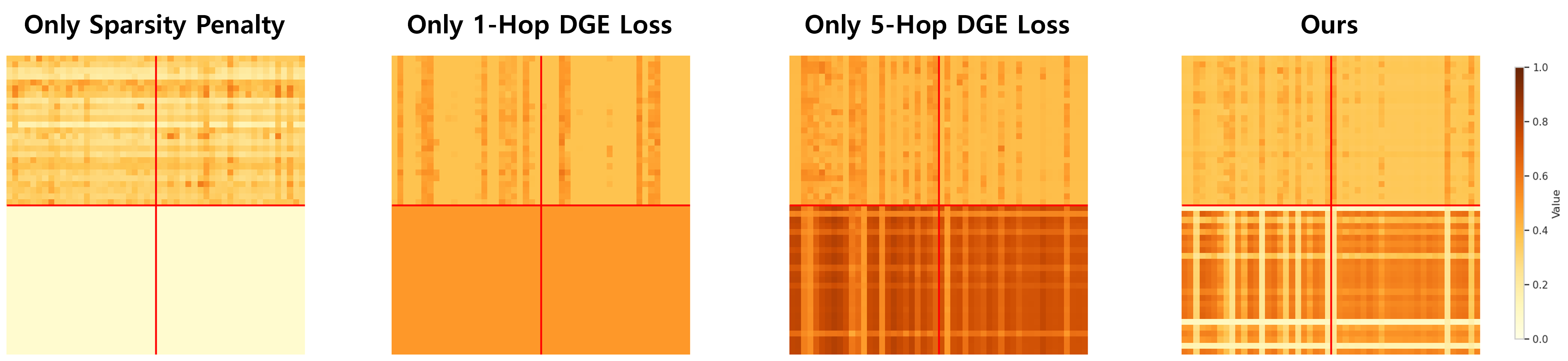}
    \caption{GRN topology analysis of GPO loss objective. Red borderlines separate perturbation and extended gene groups. The coloring scheme of the cells are based on the edge weights in each $\hat{W}$. Darker orange indicates higher probability and lighter orange indicates lower probability.}
    \label{fig:loss_analysis}
\end{figure*}

\newpage
\section{Experiment Details}

\subsection{Replogle-K562}
Each model was trained using the Adam optimizer for a total of 2,000 epochs, with a batch size of 512, a learning rate of 0.0003, and gradient clipping at a norm of 100. The data was processed with the ReplogleDataModule, and the random seed for splitting was varied across {0, 1, 2, 3, 4}. The model used was the \modelname~Model, with a latent dimension of 1,546, a single decoder layer, a mask prior probability of 0.3, and an embedding prior scale of 1. The guiding function was \modelname~CorrelatedNormalGuide, comprising 4 layers with 400 hidden units, and input normalization was applied using log standardization. The loss function utilized was \modelname~ELBOLossModule, with $\beta$ set to 0.1 and a penalty coefficient of 5.

\subsection{Replogle-RPE1}
For this experiment, each model was optimized using the Adam optimizer for 2,000 epochs, a batch size of 512, a learning rate of 0.0003, and gradient clipping with a norm of 100. Data processing was carried out using the RPE1DataModule, with the split seed varied across {0, 1, 2, 3, 4}. The model used was the \modelname~Model, with a latent dimension of 200, one decoder layer, a mask prior probability of 0.3, and an embedding prior scale of 1. The guiding function was \modelname~CorrelatedNormalGuide, consisting of 4 layers with 400 hidden units, with input normalization performed through log standardization. The loss function used was \modelname~ELBOLossModule, with $\beta$ = 0.1 and a penalty coefficient of 10.

\subsection{Adamson}
In this experiment, each model was optimized with the Adam optimizer for 2,000 epochs, a batch size of 512, a learning rate of 0.0003, and gradient clipping norm of 100. The data was processed using the AdamsonDataModule, with the split seed varied across {0, 1, 2, 3, 4}. The model employed was the \modelname~Model, with a latent dimension of 100, a single decoder layer, a mask prior probability of 0.3, and an embedding prior scale of 1. The guide was \modelname~CorrelatedNormalGuide, consisting of 4 layers with 400 hidden units, and input normalization using log standardization. The loss function was \modelname~ELBOLossModule, with $\beta$ = 0.1 and a penalty coefficient of 1.

\subsection{Implementation Details}
The baseline models were set up according to the configurations specified in the original studies. All experiments were conducted on an Ubuntu server with an NVIDIA RTX 3090Ti GPU with 24 GB of memory.

\end{appendices}
\newpage

\end{document}